\documentclass[lettersize,journal]{IEEEtran}
\usepackage{amsmath,amsfonts}
\usepackage{algorithm}
\usepackage{array}
\usepackage[caption=false,font=normalsize,labelfont=sf,textfont=sf]{subfig}
\usepackage{textcomp}
\usepackage{stfloats}
\usepackage{url}
\usepackage{verbatim}
\usepackage{graphicx}
\usepackage{cite}
\usepackage{amsmath}               
\newtheorem{assumption}{Assumption}
\usepackage{multirow}
\usepackage{amssymb}
\usepackage{makecell}
\usepackage{graphicx}
\usepackage{graphicx, subcaption}
\usepackage{algorithm}
\usepackage{algpseudocode}
\usepackage{algorithmicx}
\usepackage[table]{xcolor}
\usepackage[pagebackref,breaklinks,colorlinks,allcolors=blue]{hyperref}
\usepackage{multicol}
\usepackage{booktabs}
\usepackage{colortbl}
\usepackage{xcolor}

\definecolor{hl}{HTML}{000000}

\makeatletter
\renewcommand{\maketag@@@}[1]{\hbox{\m@th\normalsize\normalfont#1}}%
\makeatother

\begin{document}

\title{Isolating to Harness: Cross-Division Distillation \\ for Fully Unsupervised Anomaly Detection}

\author{Xinyue Liu, Jianyuan Wang, Biao Leng, Shuo Zhang
\thanks{This work was supported in part by National Natural Science Foundation of China under Grant 62402035 and in part by National Key R\&D Program of China under Grant 2024YFB4505901. \textit{(Corresponding author: Jianyuan Wang.)}}
\thanks{Xinyue Liu and Biao Leng are with the School of Computer Science and Engineering, Beihang University, Beijing 100191, China (e-mail: liuxinyue7@buaa.edu.cn; lengbiao@buaa.edu.cn).}
\thanks{Jianyuan Wang is with the Key Laboratory of Intelligent Bionic Unmanned Systems, Ministry of Education, and the School of Artificial Intelligence, University of Science and Technology Beijing, Beijing 100083, China (e-mail: wangjianyuan@ustb.edu.cn).}
\thanks{Shuo Zhang is with  School of Computer Science and Technology, Beijing Jiaotong University, Beijing 100044, China (e-mail: zhangshuo@bjtu.edu.cn).}
}

\markboth{Journal of \LaTeX\ Class Files,~Vol.~14, No.~8, August~2021}%
{Shell \MakeLowercase{\textit{et al.}}: A Sample Article Using IEEEtran.cls for IEEE Journals}

\IEEEpubid{\begin{minipage}{\textwidth}\ \\[12pt] \centering
Copyright \copyright 20xx IEEE. Personal use of this material is permitted. 
However, permission to use this material for any other purposes must \\
be obtained from the IEEE by sending an email to pubs-permissions@ieee.org.
\end{minipage}}

\maketitle

\begin{abstract}
Fully Unsupervised Anomaly Detection (FUAD) addresses the practical scenario where training data is contaminated with unlabeled anomalies.
This setting critically challenges conventional Unsupervised Anomaly Detection (UAD) methods, as they tend to misinterpret training anomalies as normal patterns, leading to false negatives. Although filtering anomalies from the training set is a common countermeasure, it inevitably discards valuable data and degrades the model's representation of normality.
To overcome this dilemma, we propose an “isolating to harness” strategy, which isolates the influence of anomalies within specialized divisions and then leverages cross-division collaboration to generate robust pseudo supervision.
We materialize this idea via a novel Cross-Division Distillation framework based on the widely studied Reverse Distillation paradigm.
CDD first partitions the data into divisions with reduced anomaly ratios to train division-specific students. It then aggregates pseudo-normal features generated by each division-specific student for samples from other data divisions to guide a global student towards a robust anomaly-free representation.
Experimental results on noisy versions of multiple AD datasets demonstrate that our method achieves significant performance improvements over the baseline.
Code is available at https://github.com/hito2448/CDD.
\end{abstract}


\section{Introduction}
\label{sec:intro}

\IEEEPARstart{I}{n} the field of industrial image anomaly detection, acquiring or predefining anomalous samples is often impractical. 
Consequently, Unsupervised Anomaly Detection (UAD), which relies only on normal samples for training, has been extensively studied~\cite{bergmann2019mvtec, bergmann2022beyond}.
To tackle the challenges of UAD task, a variety of methods are proposed, such as those based on memory banks~\cite{roth2022towards, bae2023pni}, anomaly synthesis~\cite{li2021cutpaste, zavrtanik2021draem}, and image reconstruction~\cite{bergmann2018improving, akcay2019ganomaly}. 
In recent years, UAD methods based on Knowledge Distillation (KD) have gained increasing attention~\cite{bergmann2020uninformed}. Compared to other techniques, they show greater potential in pixel-level anomaly localization. 

\begin{figure}[h]
    \centering
    \includegraphics[width=0.65\linewidth]{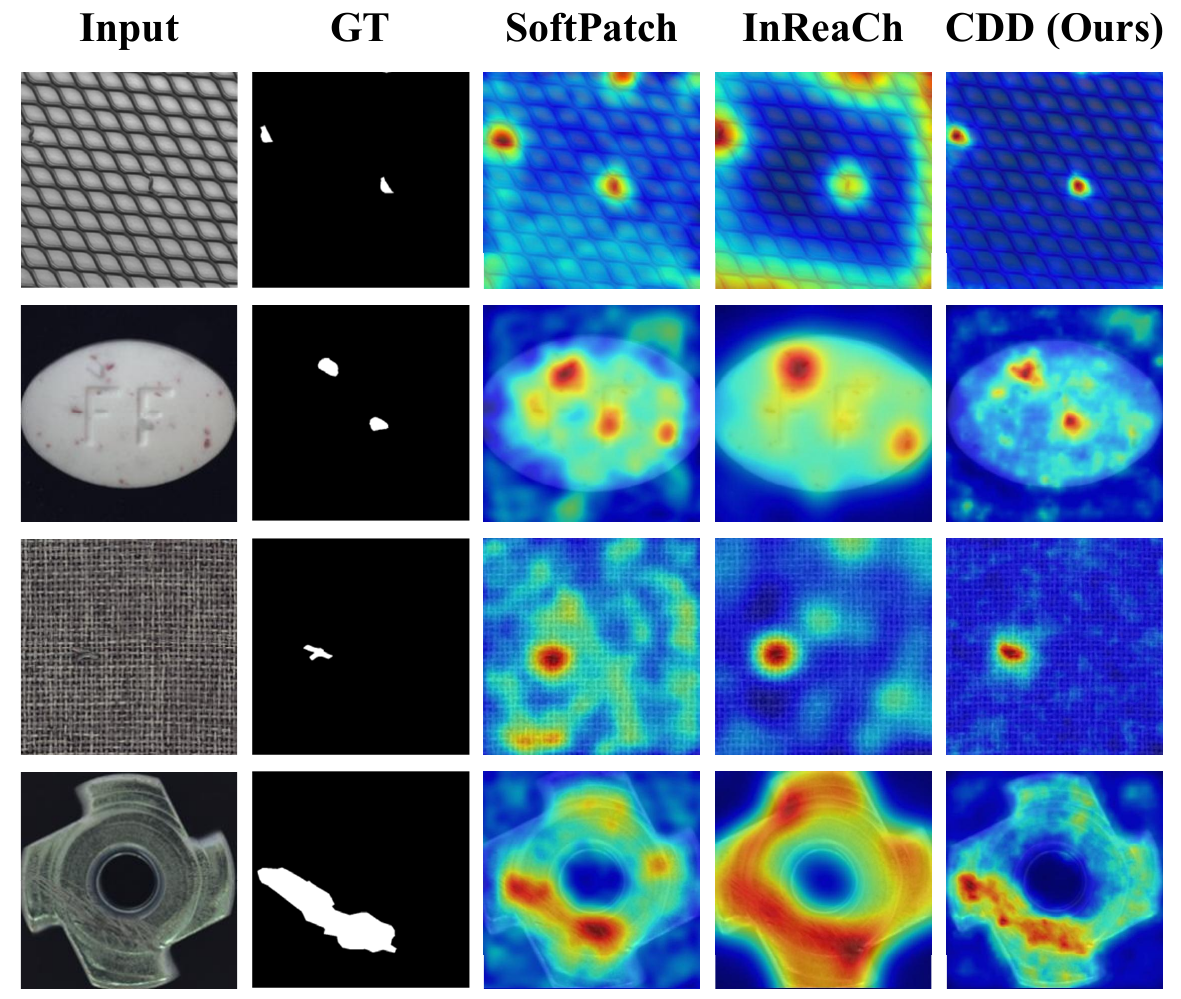}
    \caption{Qualitative comparison with other FUAD methods including SoftPatch~\cite{jiang2022softpatch} and InReaCh~\cite{mcintosh2023inter} on MVTec AD-noise-0.1~\cite{bergmann2019mvtec}. Note that throughout this paper, the notation "-noise-x" indicates that the training set is contaminated with an anomaly ratio $x$.
    }
    \label{fig:intro}
\end{figure}

\textcolor{hl}{
In real-world scenarios, however, collected unlabeled data are often imperfect and may contain anomalous, corrupted, or otherwise unreliable observations~\cite{geng2025unsupervised, yang2025improving, yan2026two}.
In industrial anomaly detection, such contamination typically appears as a small proportion of anomalous samples mixed with predominantly normal training data.
}
Relying entirely on manual data cleaning incurs high labor costs. 
This motivates the need for Fully Unsupervised Anomaly Detection (FUAD), a more practical and challenging setting where the training set may contain unlabeled anomalous samples.
However, UAD methods typically assume that the model learns only normal information and representations. If UAD methods are directly applied to the FUAD task, the model may treat anomalous samples in the training set as variations within the normal range. During testing, similar anomalies are misclassified as normal, yielding low anomaly scores as in Fig.~\ref{fig:intro} (b), increasing false negatives.

Currently, several studies have explored the FUAD task~\cite{jiang2022softpatch, mcintosh2023inter}. One common approach is to filter out suspected anomalies, retaining only normal information to convert FUAD to UAD. However, filtering strategy causes information loss: firstly, to ensure anomaly removal, a high filtering threshold excludes some normal samples, reducing training data; secondly, anomalous samples still contain normal regions, which contain usable information for training. This data loss weakens normal pattern modeling, causing normal test samples to receive high anomaly scores, resulting in false positives presented in Fig.~\ref{fig:intro} (c).

\begin{figure*}
    \centering
    \includegraphics[width=0.9\linewidth]{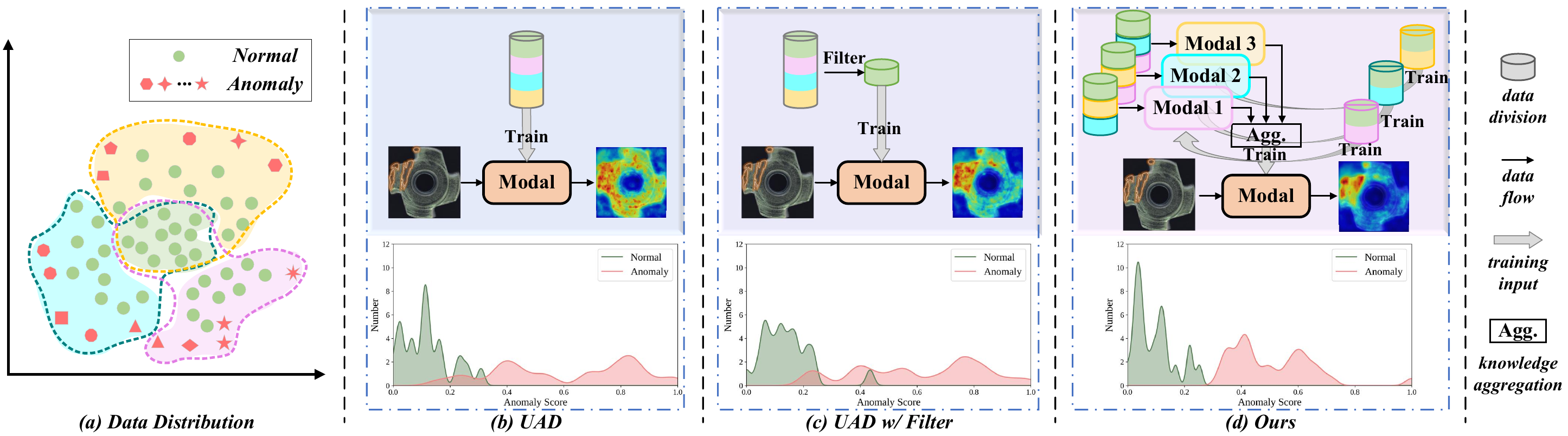}
    \caption{
     Comparison of different training \textcolor{hl}{approaches} in the FUAD setting  (MVTec AD-noise-0.4 \textit{metal\_nut}~\cite{bergmann2019mvtec}). 
     (a) Data distribution schematic. 
    (b) UAD model trained on all data, leading to false negatives during inference.
    (c) UAD model trained on the filtered training data, resulting in loss of normal patterns and false positives.
    (d) Our proposed approach (with 3 divisions), enabling full data utilization and superior anomaly detection performance.
    }
    \label{fig:intro}
\end{figure*}

\IEEEpubidadjcol
To navigate this trade-off, our objective is to leverage all available data while preventing anomalies from interfering with the learning of normal patterns.
Our approach begins with two key observations about the FUAD setting:
(1) From a probabilistic perspective, normal pixels still dominate the training set despite the presence of noise; 
(2) In terms of feature distribution, 
normal patterns are inherently more compact and stable than diverse, scattered anomalies, making them the easier-to-learn mode in the feature space.
Consequently, a standard UAD method can yield informative, albeit imperfect, initial results.
Inspired by semi-supervised learning~\cite{tarvainen2017mean, berthelot2019mixmatch}, a natural progression is to employ a naive UAD method to pre-train an auxiliary model on the entire dataset, which then generates pseudo-labels for iterative self-improvement. However, training directly on the contaminated data inevitably causes the model to reinforce its own initial mistakes, amplifying confirmation bias.

To mitigate this bias, we further discover that, due to the distributional diversity of anomalies, a UAD model still retains discrimination capability for other anomaly patterns even if it mistakenly treats a specific anomaly pattern as normal. Based on this observation, our intuition is to divide the contaminated dataset into multiple data subsets, referred to as \textit{data divisions}, and train division-specific models on each. Through intra-division isolation, the influence of anomalous samples is confined within their respective divisions, enabling each division-specific model to generate more reliable pseudo-labels for data in other divisions. By aggregating pseudo-labels from different division-specific models, a more robust overall supervision signal for anomaly detection is able to be obtained.

Following the above cross-division pseudo-supervision idea, we propose a novel \textbf{C}ross-\textbf{D}ivision \textbf{D}istillation (CDD) framework for FUAD, which is built upon the widely studied KD-based UAD paradigm Reverse Distillation (RD)~\cite{deng2022anomaly}. 
Specifically, the framework first \textbf{isolates} the influence of anomalies by partitioning the contaminated training set into multiple data divisions and training a dedicated division-specific student decoder for each. 
By leveraging the statistical diversity of anomalies across divisions,
this design encourages each division-specific student to remain ``naive" to the anomalies present in other data divisions, therefore naturally generating anomaly-free features.
Then, for samples in a given data division, we utilize division-specific students trained on other divisions to generate corresponding pseudo-normal features, guiding the training of a global student decoder. 
This process enables the global student to \textbf{harness} the entire dataset to obtain the capability of generating anomaly-free features across both normal and anomalous samples.
During Inference, the distance between the features generated by the global student decoder and the teacher encoder is used to detect and localize anomalies.

Our contributions are summarized as follows:

 \begin{itemize}
        \item We propose an ``\textit{isolating to harness}" solution for FUAD, achieving “waste-to-energy” by transforming potential anomaly interference into beneficial training signals.
        \item We 
        explore the application of the Knowledge Distillation paradigm to the Fully Unsupervised Anomaly Detection task.
        \item We propose Division-Specific Training (DST), which first performs Confidence-Guided Partition Construction to build data divisions with low anomaly probability. Then, each data division is used to train a division-specific student via Division-Specific Distillation with Regularization.
        \item Cross-Division Knowledge Aggregation (CDKA) is introduced, where division-specific students provide pseudo-normal features for out-of-division samples to train a global student integrating information across all divisions.
        \item Experimental verification on multiple datasets shows that our CDD significantly outperforms the baseline RD 
        in the FUAD setting. 
    \end{itemize}

The remainder of this paper is organized as follows. Section \ref{sec:related} reviews related work on unsupervised and fully unsupervised anomaly detection task. Section \ref{sec:motivation} elucidates the motivation behind our strategy through a visualization of the Cross-Division Distillation effect. Section \ref{sec:method} details the proposed Cross-Division Distillation (CDD) framework, including Division-Specific Training and Cross-Division Knowledge Aggregation mechanisms. Section \ref{sec:ex} presents comprehensive experimental results and ablation studies on multiple benchmarks. Finally, Section \ref{sec:conclusion} concludes the paper.

\section{Related Work}
\label{sec:related}

\subsection{Unsupervised Anomaly Detection}
Unsupervised Anomaly Detection (UAD) has been widely studied in recent years due to its ability to operate without requiring anomalous samples during training. Existing methods are broadly categorized into the following types: (1) reconstruction-based generative models~\cite{you2022unified, zhang2023unsupervised, xing2023visual, luo2025ura}, which learn to reconstruct only normal samples and identify anomalies based on reconstruction errors during inference; (2) density estimation-based methods~\cite{defard2021padim, zhou2024msflow}, which assume that normal samples follow a specific distribution in the feature space and detect deviations from this distribution; (3) synthetic anomaly-based approaches~\cite{li2021cutpaste, zavrtanik2021draem, zhang2024realnet}, which generate pseudo-anomalies using image transformations, external generators, or diffusion models to enhance the model’s ability to perceive anomalies; and (4) methods that incorporate pre-trained models and memory bank mechanisms~\cite{roth2022towards, bae2023pni, hyun2024reconpatch}, comparing the features of test samples with those of normal samples to identify anomalies. In recent years, Knowledge Distillation-based UAD methods~\cite{bergmann2020uninformed, zhou2022pull, li2024hyperbolic, salehi2021multiresolution, liu2024unistad, liu2024dualmodeling, wang2025anomaly} using the teacher-student framework have emerged as excellent methods for anomaly localization. These methods learn representations of normal regions and detect anomalies by measuring the discrepancy in features between the teacher and student networks on anomalous regions. To mitigate the student’s over-generalization to anomalies, some studies introduce heterogeneous architectures or reverse information flow, such as Reverse Distillation~\cite{deng2022anomaly} and its variants~\cite{tien2023revisiting, guo2023template, guo2024recontrast, liu2025unlocking}, which further improve anomaly detection accuracy.

\subsection{Fully Unsupervised Anomaly Detection}
Fully Unsupervised Anomaly Detection (FUAD) has attracted increasing attention, owing to its ability to operate without manual annotations and its suitability for tackling noisy training data in real-world scenarios~\cite{wang2024real}. 
\textcolor{hl}{
Beyond industrial image-based FUAD, related studies in autonomous racing and structural health monitoring have also investigated unsupervised learning from imperfect observations through anomaly filtering~\cite{geng2025unsupervised}, noise augmentation~\cite{yang2025improving}, and two-stage autoencoder reconstruction~\cite{yan2026two}. Although these studies address different data modalities and downstream tasks, they further demonstrate the practical importance of mitigating unreliable observations during unsupervised learning. 
Within industrial image-based FUAD, 
}
existing methods are categorized as follows:
(1) 
Methods based on PatchCore~\cite{roth2022towards} primarily refine the memory bank mechanism to handle noise.
SoftPatch~\cite{jiang2022softpatch} adopts a patch-level denoising strategy using noise discriminators to mitigate overconfidence. 
OPFA~\cite{zhou2024outlier} introduces an outlier probability mechanism to adaptively re-weight features, suppressing the influence of contaminated samples. 
Furthermore, TailedCore~\cite{jung2025tailedcore} extends this paradigm to tackle the specific challenge of long-tail noise distributions via few-shot sampler in the multi-class setting.
(2) InReaCh~\cite{mcintosh2023inter} builds detection models by associating high-confidence patch channels across training images. (3) FUN-AD~\cite{im2025fun} leverages nearest-neighbor distances and class homogeneity, employing an iteratively reconstructed memory bank (IRMB) to handle noisy data. However, these methods often rely on explicit memory banks, which impose storage burdens in practice. Knowledge Distillation has shown strong potential in unsupervised anomaly localization without additional storage, but its application to FUAD remains unexplored. 
\begin{figure*}
  \centering
  \includegraphics[width=0.95\linewidth]{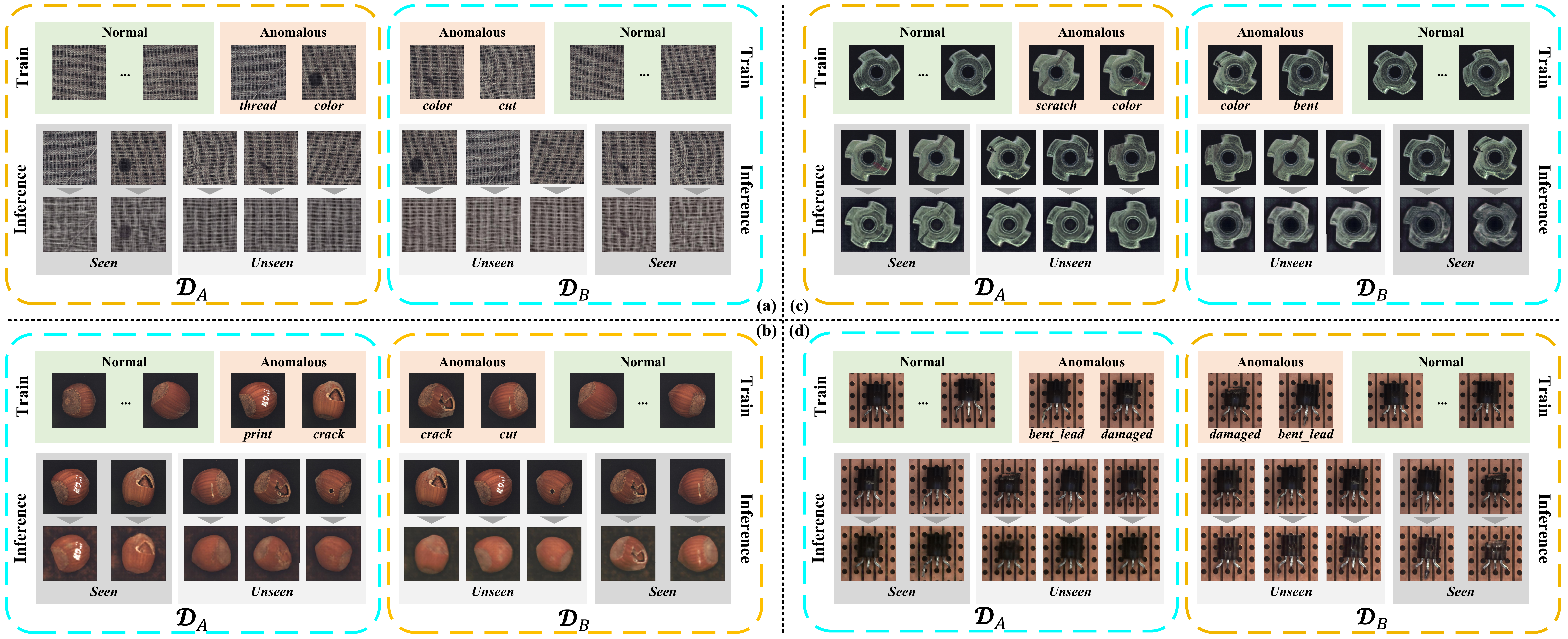}
  \caption{
  Visualization of the isolation effect on MVTec AD~\cite{bergmann2019mvtec} categories (a) \textit{carpet}, (b) \textit{hazelnut}, (c) \textit{metal\_nut}, and (d) \textit{transistor}.
\textbf{Top per division}: training images used to train the RD student decoder. 
\textbf{Bottom per division}: test images (top row) and their corresponding reconstructed student decoder features (bottom row) via the visualization decoder~\cite{you2022unified}.
  }
  \label{fig:isolate_effect}
\end{figure*}

\section{Motivation}
\label{sec:motivation}

\subsection{Rethinking Reverse Distillation for FUAD}

\noindent
\textbf{What is Reverse Distillation?}
Early KD-based AD methods typically adopt a homogeneous teacher-student framework, where the student 
only learns the teacher's representation ability on normal samples.
During inference, anomalies are detected by measuring the discrepancy between teacher and student features.
Reverse Distillation (RD)~\cite{deng2022anomaly} builds upon KD by introducing an encoder-decoder structure.
The teacher network is a frozen encoder,
while the student consists of a trainable one-class bottleneck embedding (OCBE) module \(\mathcal{B}(\cdot; \phi)\) and a trainable decoder \(\mathcal{D}_S(\cdot; \psi)\).

Let the training set be \(\mathcal{I}_{train}\). Given a training image \(I_i^{train} \in \mathcal{I}_{train}\), the teacher extracts multi-layer features \(\mathcal{F}_{\mathcal{T},i} =\mathcal{T}(I_i^{train})= \{f_{\mathcal{T},i}^l\}_{l=1}^L \), which are then reconstructed by the student network as \(\mathcal{F}_{\mathcal{S},i}  = \mathcal{S}(\mathcal{F}_{\mathcal{T},i}; \theta_\mathcal{S})= \{ f_{_\mathcal{S},i}^l\}_{l=1}^L\). The student network is denoted as \(\mathcal{S}(\cdot; \theta_\mathcal{S})\), with parameters \(\theta_\mathcal{S} = \{\phi, \psi\}\).
The training objective is to minimize the cosine distance between teacher and student features across all \(L=3\) layers on normal samples as:
\begin{small}
\begin{equation}
\cos (f_1, f_2) = \frac{ f_1 \cdot f_2 }{ \| f_1 \| \| f_2 \| },
\end{equation}
\end{small}
\begin{small}
\begin{equation}
\ell_{cos}(\mathcal{F}_\mathcal{T}, \mathcal{F}_\mathcal{S}) =  \sum_{l=1}^L \left( 1 - \cos (f_{\mathcal{T}}^l, f_{\mathcal{S}}^l) \right),
\end{equation}
\end{small}
\begin{small}
\begin{equation}
\arg\min_{\theta_\mathcal{S}} \mathbb{E}_{I_i \sim \mathcal{I}_{train}} \ell_{cos}(\mathcal{F}_{\mathcal{T},i}, \mathcal{S}(\mathcal{F}_{\mathcal{T},i}; \theta_\mathcal{S})).
\end{equation}
\end{small}

\noindent
\textbf{Why does RD Work for FUAD?}
Although Reverse Distillation (RD) is initially designed for training with only normal samples, it demonstrates strong adaptability in FUAD.
We attribute this to two key factors:

(1) \textbf{Probability Perspective} - Dominance of Normal Samples: \\
In industrial scenarios, normal samples are much more common than anomalies, which results in low proportion of anomalous images in the training set. Moreover, anomalies typically occupy only a small region within an image. 
Consequently, the student network, driven by the dominance of normal samples, primarily learns to represent normal features, while the sparsity of anomalies limits their impact on the optimization process.

(2) \textbf{Distribution Perspective} - Concentrated Normal vs. Diverse Anomaly: \\
Normal samples exhibit compact and consistent feature patterns, while anomalous samples are diverse and scattered.
Thus, it is difficult for the student to generalize learned anomalous features.
Even if the student learns representations of anomalies in the training set, it tends to generate normal features for unseen test anomalies.
When encountering unseen anomalies during inference, the student's output deviates significantly from the teacher’s, forming the basis for anomaly detection.

\noindent
\textbf{Challenges of Applying RD to FUAD.}
In FUAD task, the training set naturally includes a certain proportion of anomalous samples. 
If highly similar or recurring anomalies appear repeatedly during training, the student may learn to
reconstruct their teacher features.
This results in poor discrimination against similar anomalies during testing.
Therefore, the key challenge in applying RD to FUAD is \textit{how to prevent the student from modeling seen anomaly features during training, to ensure that it generates anomaly-free features for all other possible anomalous samples (even for samples seen during training)}.

\noindent
\textbf{The Dilemma of Data Utilization vs. Noise Suppression.}
To mitigate the risk of anomaly modeling in the FUAD setting, a straightforward countermeasure adopted by existing FUAD methods~\cite{jiang2022softpatch, zhou2024outlier} is to explicitly filter out suspected anomalies. 
However, this passive filtering strategy is inherently suboptimal for data-driven paradigms like RD.
On one hand, data-driven methods thrive on the diversity and volume of data. Aggressive filtering inevitably reduces the effective sample size and discards valid normal information embedded within anomalous images such as normal backgrounds, degrading the density of the learned normal distribution. 
On the other hand, simply hiding anomalies prevents the student from actively learning to be robust against noise. 
Our intuition is that these noisy anomalous samples, rather than being mere interference, serve as valuable resources. 
If properly managed, they force the student network to generalize the normality modeling capability and learn to generate anomaly-free features even when the input contains seen anomaly patterns.

To achieve the aforementioned goal of utilizing all data while distinguishing anomalies from normality, our key insight is to isolate the influence of different anomalies rather than removing them. To realize this, we propose Cross-Division Distillation (CDD) based on an ``\textit{isolating to harness}'' strategy, which partitions the contaminated training data into multiple data divisions and training division-specific RD students that are sensitive only to anomalies within their own division. 
In this way, each division-specific student naturally treats anomalies from other divisions as out-of-distribution and generates pseudo-normal features for them.

\begin{figure*}
  \centering
  \includegraphics[width=0.95\linewidth]{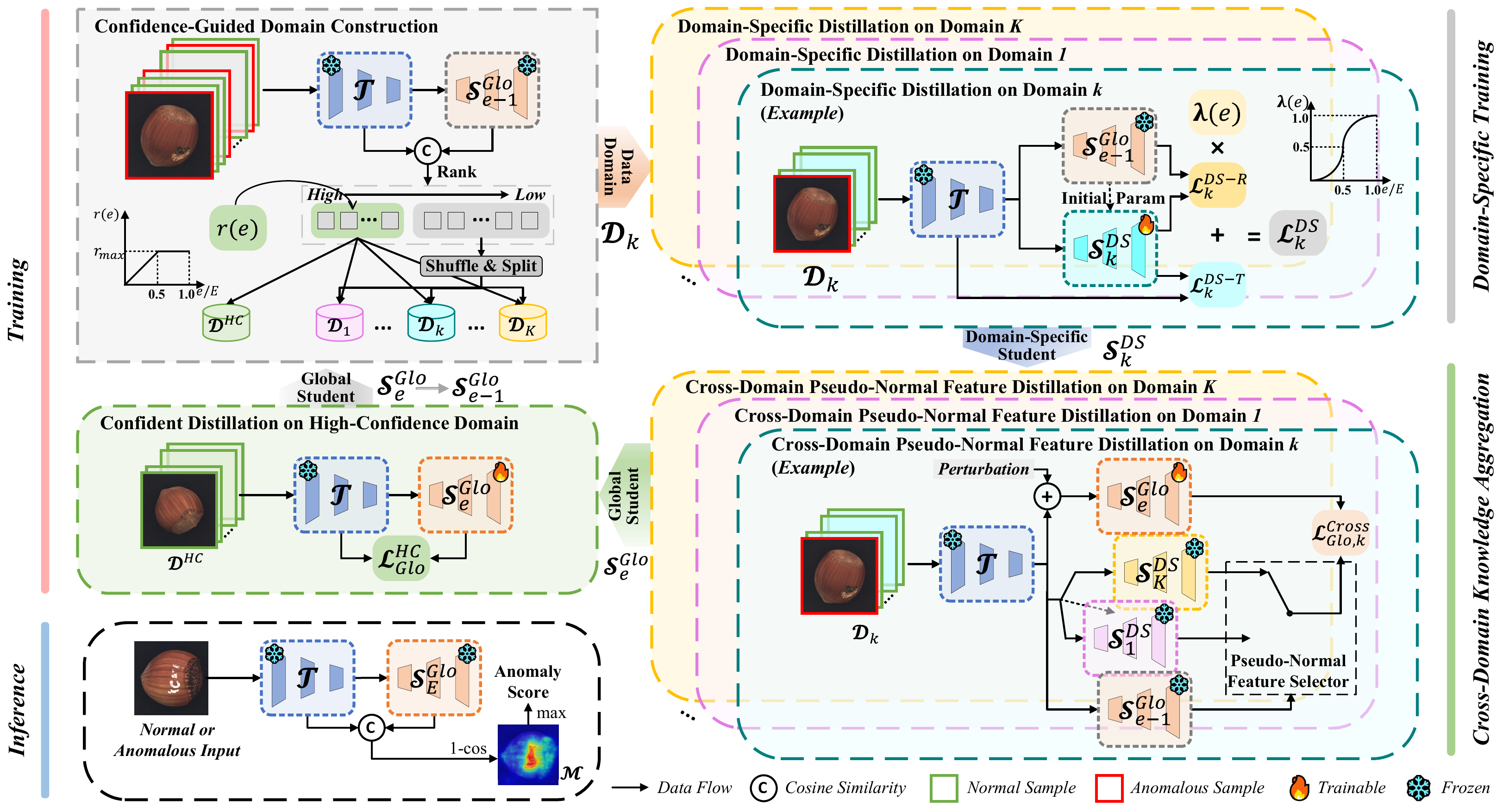}
  \caption{
  Overall framework of our proposed CDD.
  \textbf{Top \& bottom right}: training pipeline with Division-Specific Training and Cross-Division Knowledge Aggregation. \textbf{Bottom left}: inference using the global student and teacher.
  }
  \label{fig_overview}
\end{figure*}

\subsection{Design Rationale of Cross-Division Distillation}
\label{sec:theory}

In the previous section, we intuitively analyze the feasibility of RD for FUAD from both probability and distribution perspectives. 
In this section, we summarize these intuitions into two assumptions.
They not only elucidate the underlying mechanism enabling data-driven methods like RD to function in FUAD setting but also constitute the 
\textcolor{hl}{design rationale}
for our proposed CDD framework.
\begingroup
\color{hl}
Detailed analyses from the empirical-risk and feature-distribution perspectives are provided in the \textit{Supplementary Material}.

\begin{assumption}[Normal-Dominant Convergence]
\label{a1}
Because normal samples and normal regions dominate contaminated industrial training data and exhibit relatively consistent feature patterns, the optimization of an RD student is primarily governed by normal reconstruction. Sparse and diverse anomalous features provide less consistent learning signals and are therefore less likely to be stably reconstructed.
\end{assumption}

\textbf{Implication for CDD.}
Assumption~\ref{a1} provides a mechanism-level explanation for the noise tolerance of RD training
and serves as the guiding principle for our \textit{Confidence-Guided Division Construction} in Section \ref{sec:construction}.
While standard RD relies passively on the natural rarity of anomalies, our method actively amplifies this effect.
Specifically, sharing high-confidence samples across all divisions provides sufficient and consistent normal support for each division-specific student, while distributing the remaining low-confidence samples limits the repeated influence of individual anomalous samples. In this way, each student is encouraged to preserve a stable normal reconstruction ability without discarding the useful normal regions contained in contaminated images.

\begin{assumption}[Limited Cross-Sample Anomaly Generalization]
\label{a2}
Unlike normal features, which exhibit relatively consistent characteristics across samples, anomalous features often contain sample-dependent variations. Consequently, even if a student network overfits to reconstruct the anomalous features of a particular training sample, this sample-specific reconstruction ability cannot be reliably generalized to another unseen anomalous sample.
\end{assumption}

\textbf{Implication for CDD.}
Assumption~\ref{a2} motivates training different students on different data divisions and exploiting their outputs through the \textit{Cross-Division Knowledge Aggregation} in Section~\ref{sec:CDKA}. For a sample assigned to one division, students trained on the other divisions have not directly observed that sample and are therefore less likely to accurately reproduce the anomalous features specific to it. Meanwhile, supported by the normal-dominant learning described in Assumption~\ref{a1}, these students retain the ability to reconstruct its shared normal features. Their outputs thus provide candidate features in which the influence of the target anomaly is suppressed while the normal information is preserved.

\textbf{Remark.}
This assumption is more likely to hold when anomalies exhibit substantial instance-level variations, but may be weakened by repeatedly occurring highly similar or near-duplicate anomalies. The effect of cross-division anomaly similarity, which may become more pronounced at higher contamination ratios, is further examined in Section~\ref{sec:anomaly_similarity}.

\endgroup

\subsection{Empirical Verification of Cross-Division Distillation}
\label{sec:empirical_motivation}

To empirically validate the effectiveness of our proposed Cross-Division Distillation, we investigate whether the student network trained on one data division can generate candidate pseudo-normal features for anomalies in another division.

\textbf{Setup.}
We conduct a pilot experiment on the MVTec AD dataset~\cite{bergmann2019mvtec}. For a specific category, we partition the contaminated training data into two disjoint divisions, $\mathcal{D}_A$ and $\mathcal{D}_B$.
Consequently, while $\mathcal{D}_A$ and $\mathcal{D}_B$ consist of distinct anomaly samples, they naturally share anomalies of the same semantic label (e.g., both contain  anomalous images with label ``\textit{color}").
Two independent RD students, $\mathcal{S}_A$ and $\mathcal{S}_B$, are trained exclusively on $\mathcal{D}_A$ and $\mathcal{D}_B$, respectively.
We then visualize features generated by $\mathcal{S}_A$ and $\mathcal{S}_B$ using the corresponding visualization decoders~\cite{you2022unified} on both seen (intra-division) and unseen (cross-division) anomalies.

\textbf{Observation.}
As depicted in Fig. \ref{fig:isolate_effect}, a clear asymmetric pattern emerges:
\begin{itemize}
\item For \textbf{seen anomalies} (e.g., $\mathcal{S}_A$ on anomalies from $\mathcal{D}_A$), the student reconstructs teacher-like features, indicating it has learned division-specific anomalies.
\item For \textbf{unseen anomalies} (e.g., $\mathcal{S}_A$ on anomalies from $\mathcal{D}_B$), the student fails to reconstruct them well, and intends to produces features that closely resemble its learned ``normal" patterns. This pseudo-normal generation occurs even for samples in $\mathcal{D}_B$ that share the same anomaly label as those in $\mathcal{D}_A$.
\end{itemize}

\textbf{Implication for CDD.}
\textcolor{hl}{
The observed behavior provides empirical support for both design premises. First, the tendency of the student to produce normal-like features for unobserved anomalies is consistent with the normal-dominant learning described in Assumption~\ref{a1}. Second, the persistence of this behavior despite the presence of same-label anomalies in the training division supports the limited cross-sample transferability described in Assumption~\ref{a2}. 
These observations suggest that sample-level isolation prevents a cross-division student from directly memorizing the target anomaly, while preserving its ability to reconstruct the shared normal content. Therefore, students trained on other divisions provide candidate pseudo-normal features that are less affected by the memorization of the target sample. CDD further aggregates and selects among these candidates to construct more reliable pseudo-normal supervision for the global student.
}

\section{Method} \label{sec:method}

\subsection{Overview}

\subsubsection{Problem Definition}
In FUAD, we denote the training set as \( \mathcal{I}_\text{train} = \{I_i^\text{train}\}_{i=1}^{N} \), where each image 
\( I_i^\text{train} \in \{\mathcal{N}, \mathcal{A}\} \) 
is unlabeled and may be normal or anomalous. 
The test set \( \mathcal{I}_\text{test} = \{I_j^\text{test}\}_{j=1}^{M} \) comprises both normal and anomalous images, with normal samples following the same distribution as \( \mathcal{I}_\text{train} \). The objective is to learn the distribution of normal samples from \( \mathcal{I}_\text{train} \) to detect anomalies in \( \mathcal{I}_\text{test} \).

\subsubsection{Overall Training Process}
Fig.~\ref{fig_overview} illustrates the training process of each epoch (top and lower right) and the inference process (lower left). All teacher and student networks follow the design of RD. The teacher is an pre-trained encoder such as WideResNet-50 \cite{zagoruyko2016wide}.
Each student includes an OCBE module (or a Noisy Bottleneck module) and a decoder.

Each training epoch consists of two stages: Division-Specific Training and Cross-Division Knowledge Aggregation.
In the first stage, we propose Confidence-Guided Division Construction to extract high-confidence normal samples from the original training set and use them as the intersection between multiple data divisions. In this way, each division has a reduced anomaly ratio compared to the full dataset.
Then, we train a division-specific student for each data division using Division-Specific Distillation with Regularization, which focuses on modeling teacher-like normal features in its local division.
The second stage Cross-Division Knowledge Aggregation mainly explains how to use the division-specific students obtained in the first stage to train a global student that reconstructs normal features on all samples. 
For anomalous samples in a specific division \(k\) , division-specific students that are not trained on division \(k\) still generates normal-like features. We use these features as pseudo-normal supervision signals to perform Cross-Division Pseudo-Normal Feature Distillation for the global student. After that, we further distill the global student using the teacher on high-confidence normal samples, enabling it to effectively learn the reliable reconstruction of normal patterns.

\subsubsection{Overall Inference Process}

The lower left part of Fig.~\ref{fig_overview} depicts the inference process.
During inference, for each image \(I_j^{test} \in \mathcal{I}_{test}\), cosine distances across multi-layer features generated by the teacher \(\mathcal{T}\) and the global student trained for \(E\) epochs \(\mathcal{S}^{Glo}_E\) are fused to generate a pixel-level anomaly map \( \mathcal{M}\), whose maximum value serves as the image-level anomaly score \( s\):
\begin{small}
\begin{equation}
\mathcal{M}(h,w) = \sum_{l=1}^L \left( 1 - \cos (f_{\mathcal{T}}^l(h,w), f_{\mathcal{S}^{Glo}_E}^l(h,w)) \right),
\end{equation}
\end{small}
\begin{small}
\begin{equation}
s=\max(\mathcal{M}).
\end{equation}
\end{small}

\subsection{Division-Specific Training}
\label{sec:DST}

\subsubsection{Confidence-Guided Division Construction}
\label{sec:construction}

The foundation of our framework is to partition the training set into divisions for training division-specific students. While our core insight in Section~\ref{sec:empirical_motivation} shows that division-specific students naturally handle anomalies from other divisions, we argue that their ability to learn a high-quality normal representation remains crucial. If a data division contains an excessively high concentration of challenging or anomalous samples, its dedicated student may learn a biased or distorted notion of normality, which could propagate errors during cross-division aggregation.

To proactively create a more favorable learning environment for each division-specific student, we enhance the naive equal partitioning with a confidence-guided strategy. Specifically, we inject a portion of high-confidence normal samples into each division. This ensures that:
(1) The anomaly ratio within each data division is actively lowered, reducing the immediate interference of anomalies and providing a stable foundation for learning normal patterns.
(2) The core normal distribution in each division remains representative of the overall data, preventing the students from overfitting to small, noisy subsets.

We use the features output by the global student of the previous epoch \( \mathcal{S}^{Glo}_{e-1} \) as the basis for confidence evaluation. For each training sample \(I_i\), the average cosine similarity between teacher features \(f_{\mathcal{T}}\) and global student features \(f_{\mathcal{S}^G_{e-1}}\) across \(L\) layers is calculated to obtain the corresponding \(\mathrm{Conf}_i\):
\begin{small}
\begin{equation}
\begin{split}
\mathrm{Conf}_i = \sum_{l=1}^L \left\{ \frac{1}{H_lW_l} \sum_{h=1}^{H_l}\sum_{w=1}^{W_l} \cos ( f_{\mathcal{T},i}^l(h,w), 
f_{\mathcal{S}^{Glo}_{e-1},i}^l(h,w)) \right\}.
\end{split}
\end{equation}
\end{small}
All samples are sorted by confidence in descending order. The top \(r(e)\) samples form the high-confidence set \(\mathcal{D}^{HC}\). The \textcolor{hl}{high-confidence ratio} \(r(e)\) increases with training, up to 50\%. 
\textcolor{hl}{
Let \(e\) denote the current epoch, \(E\) the total number of epochs, and \(e_s\) the activation epoch of confidence guidance. The high-confidence ratio is defined as
\begin{small}
\begin{equation}
r(e)=\min\left(\max\left(\frac{e-e_s}{E-e_s},\,0\right),\,r_{\max}\right),
\label{eq:hc_ratio}
\end{equation}
\end{small}
where \(r_{\max}=0.5\) and \(e_s=0\). Before confidence guidance is activated, \(r(0)=0\), and thus the high-confidence set \(\mathcal{D}^{HC}\) remains empty, avoiding the use of unreliable confidence estimates from the untrained global student at the first epoch. As training proceeds, \(r(e)\) progressively increases and is capped at \(r_{\max}\), allowing an increasing number of samples with high normal confidence to be shared across the data divisions.
}

The remaining low-confidence samples are randomly and evenly divided into \(K\) subsets, denoted \({\mathcal{D}^{LC}_k},{k=1,\dots,K}\). 
By combining \(\mathcal{D}^{HC}\) and \(\mathcal{D}^{LC}_k\), each division is expressed as
\begin{small}
\begin{equation}
\mathcal{D}_k = \mathcal{D}^{HC} \cup \mathcal{D}^{LC}_k, \quad k = 1, \dots, K.
\end{equation}
\end{small}

\subsubsection{Division-Specific Distillation with Regularization}
After division construction, we train a corresponding division-specific student \(\mathcal{S}_k^{DS}\) for the \(k\)-th data division, who learns to reconstruct the features of samples within its corresponding division. 
The initial parameters of each division-specific student are inherited from the global student of the previous epoch \(\mathcal{S}^{Glo}_{e-1}\).
This training process of \(\mathcal{S}_k^{DS}\) follows the basic framework of Reverse Distillation, which minimizes the cosine distance between the features generated by the student \(\mathcal{F}_{\mathcal{S}_k^{DS}}\) and the features of the teacher \(\mathcal{F}_{\mathcal{T}}\). 
In this way, the division-specific students are able to model the teacher's feature representation ability of data in their local division.

However, our division construction is not flawless. Despite our efforts, highly similar anomalies might appear across different divisions.
In such cases, if division-specific students are trained solely to mimic the frozen teacher, they risk learning to reconstruct these similar anomalous features.
This can reduce the quality of the cross-division candidate features and weaken the subsequent Cross-Division Knowledge Aggregation phase.
Specifically, when a division-specific student has learned to reconstruct anomalies similar to the target sample, it may output features close to the anomalous teacher features rather than more normal-like features when serving as a pseudo-label generator for other divisions.

To further tackle this problem, 
we introduce the global student from the previous epoch \(\mathcal{S}^{Glo}_{e-1}\) to provide a regularization signal, complementary to the primary teacher supervision.
The intuition is that the global student, benefiting from knowledge aggregation, is generally less sensitive to anomaly features than individual division-specific students. 
By encouraging consistency with the global student's output features, we apply a soft constraint that gently steers division-specific students away from accurately reconstructing anomaly features, biasing them towards the global normal consensus. 
Furthermore, acknowledging that the global student might be under-trained in the initial stages, we implement a dynamic scheduling strategy where the regularization weight \(\lambda(e)\) increases gradually. 
This ensures that the regularization strength grows with the reliability of the global student.

The loss \(\mathcal{L}^{DS}_k\) used to train each division-specific student \(\mathcal{S}_k^{DS}\) combines two terms: the primary distillation loss (from the teacher) and the regularization loss (from the global student), which is expressed as
\begin{small}
\begin{equation}
\begin{split}
\mathcal{L}^{DS}_k = \mathbb{E}_{I_i \sim \mathcal{D}_{k}} (& \underbrace{\ell_{cos}(\mathcal{F}_{\mathcal{T},i}, \mathcal{F}_{\mathcal{S}_k^{DS},i})}_{\text{Teacher Guidance}\  \mathcal{L}^{DS-T}_k} \\
 & + \lambda(e) \cdot\underbrace{ \ell_{cos}(\mathcal{F}_{\mathcal{S}^{Glo}_{e-1},i}, \mathcal{F}_{\mathcal{S}_k^{DS},i})}_{\text{Regularization}\ \mathcal{L}^{DS-R}_k}),
\end{split}
\end{equation}
\end{small}
where \(\lambda(e)\) is a dynamic increasing coefficient that adjusts the regularization strength over the training epochs. It is controlled using an S-shaped scheduling function with \(p=4.0\) as
\begin{small}
\begin{equation}
\lambda(e) = \frac{(e/E)^p}{(e/E)^p + (1-e/E)^p}.
\end{equation}
\end{small}
\textcolor{hl}{
Since \(\lambda(e)\) is close to zero during the early training stage,
the division-specific students are mainly optimized by teacher guidance,
while the regularization from the previous global student is introduced
progressively.
}

\subsection{Cross-Division Knowledge Aggregation}
\label{sec:CDKA}

\subsubsection{Cross-Division Pseudo-Normal Feature Distillation}

Due to the high consistency of normal samples across data divisions, division-specific students reconstruct reliable normal features on normal samples in all divisions. 
However, the diversity of anomalies prevents division-specific students from generalizing to out-of-division anomalies, even if they learn the reconstruction of anomaly features in local data divisions. 
Following this idea, we propose using division-specific students to generate pseudo-normal features for out-of-division samples. 

This provides a foundation for training the global student: for each sample, we obtain multiple candidate pseudo-normal features from the division-specific students of other divisions. The key challenge then becomes how to aggregate these candidates. Since similar anomalies might appear across data divisions, some students might still generate anomalous features for a given sample. To prevent this contamination, 
we design a \textit{Global-Guided Pseudo-Normal Feature Selection} strategy, which identifies and selects the most reliable pseudo-normal feature from the candidates.

Specifically, 
we employ the global student as a guide to select the most reliable pseudo-normal feature from the candidates generated by division-specific students of other divisions.
The core motivation is that the global student, effectively aggregating information from the entire dataset, serves as the most neutral representation of normality. Therefore, it functions as a stable anchor to evaluate the reliability of candidate features.
In the implementation, we achieve pseudo-normal feature selection by eliminating outlier features that are more likely to be abnormal features from the output features of division-specific students with the help of the global student from the previous epoch \( \mathcal{S}^{Glo}_{e-1} \).

For a sample \(I_i\)  from data division \(\mathcal{D}_k\), we first extract features \(\mathcal{F}_{\mathcal{S}_h^{DS},i}= \{f_{\mathcal{S}_h^{DS},i}^l\}_{l=1}^L, h=\{1,\dots,K\} \setminus k\) using the division-specific students from divisions \(\mathcal{D}_h, h=\{1,\dots,K\} \setminus k\), and obtain the reference features \(\mathcal{F}_{\mathcal{S}^{Glo}_{e-1},i}= \{f_{\mathcal{S}^{Glo}_{e-1},i}^l\}_{l=1}^L\) the global student from the previous epoch. We construct an affinity vector \(\text{Aff}_i \in \mathbb{R}^{(K-1) }\), where each element measures the cosine similarity between the flattened features of each student and the global student:
\begin{small}
\begin{equation}
\text{Aff}_i = \sum_{l=1}^L \cos(f_{\mathcal{S}_h^{DS},i}^l, f_{\mathcal{S}^{Glo}_{e-1},i}^l), \quad  h=\{1,\dots,K\} \setminus k.
\end{equation}
\end{small}
The pseudo-normal feature for the training sample is then selected as the one with the highest similarity:
\begin{small}
\begin{equation}
\mathcal{F}_{\text{pseudo}, i} = \mathcal{F}_{\mathcal{S}_{h^*}^{DS},i}, \quad h^* = \arg\max_h \text{Aff}_i(h).
\end{equation}
\end{small}

\begingroup \color{hl}
Although the global-guided selection improves the reliability of the pseudo-normal supervision, \(\mathcal{F}_{\mathrm{pseudo},i}\) is dynamically generated by a division-specific student rather than obtained from ground-truth normal data and may therefore still contain transient or sample-specific reconstruction errors. Inspired by the central principle of weak-to-strong consistency learning~\cite{sohn2020fixmatch}, we use the pseudo-normal feature generated from the unperturbed cross-division branch as the reference target and train the global student using a perturbed feature-space view. Since the global student operates on teacher features rather than raw images, Gaussian feature jittering is adopted as the perturbation operator.

Specifically, for each feature level \(l\), we randomly sample a perturbation scale \(\sigma^l\) and Gaussian noise \(\delta_i^l\) as
\begin{small}
\begin{equation}
\sigma^l\sim\mathcal{U}(0,\sigma_{\max}), \qquad \delta_i^l\sim\mathcal{N}\left(\mathbf{0},(\sigma^l)^2\mathbf{I}\right).
\end{equation}
\end{small}
Here, \(\sigma_{\max}\) denotes the maximum perturbation strength and is set to \(0.2\) in all experiments. 
The perturbed teacher features are then fed into the global student:
\begin{small}
\begin{equation}
\mathcal{F}_{\mathcal{S}^{Glo}_e,i}^{*}=\mathcal{S}\left(\left\{f_{\mathcal{T},i}^l+\delta_i^l\right\}_{l=1}^L;\theta_{\mathcal{S}^{Glo}_e}\right).
\end{equation}
\end{small}
By requiring feature views with different perturbation strengths to match the pseudo-normal target generated in the current epoch, this strategy encourages the global student to produce locally consistent predictions around the teacher feature and reduces its sensitivity to inaccuracies in the pseudo-normal supervision. The pseudo-normal target is regenerated in every epoch, and thus the perturbation does not constrain the global student to a fixed pseudo target throughout training. Moreover, randomly sampling the perturbation scale from \(0\) to \(\sigma_{\max}\) exposes the global student to a continuous range of feature-space perturbations and avoids dependence on a single fixed noise level.
\endgroup

The loss of Cross-Division Pseudo-Normal Feature Distillation \(\mathcal{L}_{Glo}^{Cross}\) is then defined as:
\begin{small}
\begin{equation}
\mathcal{L}_{Glo}^{Cross} = \sum_{k=1}^K \mathbb{E}_{I_i \sim \mathcal{D}_{k}} \ell_{cos}(\mathcal{F}_{\text{pseudo},i}, \mathcal{F}_{\mathcal{S}^{Glo}_e,i}^*).
\end{equation}
\end{small}

\subsubsection{Confident Distillation on High-Confidence Divisions}

To further anchor the global student in reliable normality, we incorporate the high-confidence set \(\mathcal{D}^{HC}\). 
This step serves as a critical countermeasure against the confirmation bias that may arise from cross-division training. Since the pseudo-normal features used in the previous stage are essentially student-generated predictions, relying solely on them may introduce unstable supervision.

In contrast, $\mathcal{D}^{HC}$ consists of samples that are statistically most likely to be normal. By enforcing direct consistency with the teacher's features on these high-confidence normal samples, we provide a ``correction signal" that realigns the global student's representation with the authentic normal manifold. This ensures that the global student remains grounded in real teacher supervision while benefiting from the extensive cross-division pseudo-supervision.

For the samples in \(\mathcal{D}^{HC}\), we use the teacher's features as direct supervision to refine the global student’s modeling ability of normal patterns:
\begin{small}
\begin{equation}
\mathcal{L}_{Glo}^{HC} = \mathbb{E}_{I_i \sim \mathcal{D}^{HC}} \ell_{cos}(\mathcal{F}_{\mathcal{T},i}, \mathcal{F}_{\mathcal{S}^{Glo}_e,i}).
\end{equation}
\end{small}

\section{Experiments}
\label{sec:ex}

\subsection{Experimental Setup}  \label{sec:exsetup}

\subsubsection{Datasets}
We conduct experiments on three widely-used datasets: MVTec AD~\cite{bergmann2019mvtec}, VisA~\cite{zou2022spot}, and MVTec 3D-AD~\cite{bergmann2021mvtec} (RGB version).
\textbf{MVTec AD} contains 5 texture and 10 object categories, with 3629 normal training images and 1725 test images including both normal and anomalous samples.
\textbf{VisA} consists of 12 categories grouped into three types: Complex Structure, Multiple Instances, and Single Instance. It includes 9621 normal and 1200 anomalous images. 
\textbf{MVTec 3D-AD} comprises 10 distinct industrial object categories, containing 2656 normal training samples and 1197 test samples, where we exclusively utilize the RGB images for our experiments.

Since both datasets are originally designed for unsupervised anomaly detection, we adapt them to the FUAD setting following SoftPatch~\cite{jiang2022softpatch}. 
Specifically, normal training images are retained while a controlled portion of anomalous test samples is injected into the training set with a predefined ratio \( R_{\text{noise}}\).
We evaluate under two settings: 
(1) \textit{No overlap} setting, where injected anomalous samples are removed from the test set; and 
(2) \textit{Overlap} setting, where these anomalies remain in the test set, providing a comprehensive assessment of the model's ability to detect all anomalies, including those seen during training.
\textcolor{hl}{
The training pipeline, model configuration, and inference procedure remain identical under the two settings.
}

\subsubsection{Implementation Details}
To demonstrate the effectiveness of our proposed method, we instantiate CDD on two distinct baselines: a CNN-based version denoted as \textbf{CDD$^*$} (based on the original RD~\cite{deng2022anomaly}) and a Transformer-based version denoted as \textbf{CDD$^\dagger$} (based on Dinomaly~\cite{guo2025dinomaly}, which is designed following RD paradigm).

\textbf{Network Architecture.}
For CDD$^*$, following the standard RD~\cite{deng2022anomaly}, we utilize a WideResNet-50~\cite{zagoruyko2016wide} pre-trained on ImageNet as the backbone.
For CDD$^\dagger$, we employ a ViT-Base~\cite{dosovitskiy2020image} by DINOv2-R~\cite{darcet2023vision} as the encoder, consistent with the implementation of Dinomaly~\cite{guo2025dinomaly}.

\textbf{Input Resolution.}
Image preprocessing strategies differ to align with the respective baselines:
\begin{itemize}
    \item For CDD$^*$, following standard protocols in SoftPatch~\cite{jiang2022softpatch}, images are resized to $256 \times 256$ and center-cropped to $224 \times 224$.
    \item For CDD$^\dagger$, adhering to the training configuration of Dinomaly~\cite{guo2025dinomaly}, images are resized to $448 \times 448$ and center-cropped to $392 \times 392$ for both training and inference. 
\end{itemize}

\textbf{Training Configurations.}
We train a separate model for each category.
Both variants are optimized using the Adam optimizer with a learning rate of 0.005. All division-specific students and the global student are trained for 200 epochs. 
\textcolor{hl}{
For all main experiments on three datasets, we uniformly adopt the same dynamic division schedule \(K=\{2,3,3,2\}\) without dataset-specific re-tuning. 
Specifically, \(K\) is set to \(2\) during the first and final quarters of training and to \(3\) during the middle half.
}

\textbf{Evaluation Configurations.}
To ensure a fair comparison, the anomaly maps generated by all compared methods including CDD$^\dagger$ are bilinearly interpolated to $224 \times 224$ before evaluation.
To smooth the final anomaly maps, we apply Gaussian filtering with $\sigma = 4$.
All experiments are conducted on a single Nvidia GeForce RTX 3090 GPU.

\subsubsection{Evaluation Metrics}
We use the area under the ROC curve (AUC) at both image and pixel levels, denoted as I-AUC and P-AUC, to evaluate anomaly detection and localization performance. 
I-AUC measures the ability to distinguish anomalous images, while P-AUC evaluates pixel-wise localization accuracy. 
The per-region-overlap (PRO) metric is also reported to better evaluate the localization performance of anomalies with small sizes.

\begin{table*}
  \caption{Anomaly detection and localization results in \textit{No Overlap}  and \textit{Overlap} settings on MVTec AD-noise-0.1~\cite{bergmann2019mvtec} with the best in bold
  and the second best underlined.
  For fair comparison, all baseline results are reproduced in the same experimental environment. 
  }
  \label{tab_all_results:mvtec}
  \centering
  \resizebox{0.9\linewidth}{!}{
  \begin{tabular}{c|c|ccc|ccccc}
    \toprule
   \multicolumn{2}{c|}{Type} & \multicolumn{3}{c|}{Unsupervised} &  \multicolumn{5}{c}{Fully Unsupervised} \\
    \midrule
    Setting & Metric  & RD~\cite{deng2022anomaly}  & RD++~\cite{tien2023revisiting}  & Dinomaly~\cite{guo2025dinomaly} & SoftPatch~\cite{jiang2022softpatch} & InReaCh~\cite{mcintosh2023inter} & FUN-AD~\cite{im2025fun}  & \cellcolor{blue!10}CDD$^*$ (Ours) & \cellcolor{blue!10}CDD$^\dagger$ (Ours)\\
    \midrule
    \multirow{4}{*}{\textit{No Overlap}} & I-AUC & 0.972 & 0.978 & \underline{0.989} & 0.985 & 0.917 & 0.975 & \cellcolor{blue!10}0.984 & \cellcolor{blue!10}\textbf{0.992}\\
                                         & P-AUC & 0.957 & 0.961 & 0.971 & 0.975 & 0.971 & 0.980 & \cellcolor{blue!10}\underline{0.981} & \cellcolor{blue!10}\textbf{0.986}\\
                                         & PRO   & 0.916 & 0.921 & \underline{0.931} & 0.930 & 0.877 & 0.869 & \cellcolor{blue!10}0.930 & \cellcolor{blue!10}\textbf{0.941}\\
                                    \cmidrule(r){2-10}
                                         & Avg.  & 0.948 & 0.953 & 0.964 & 0.963 & 0.922 & 0.942 & \cellcolor{blue!10}\underline{0.965} & \cellcolor{blue!10}\textbf{0.973}\\
                                    \midrule
    \multirow{4}{*}{\textit{Overlap}}    & I-AUC & 0.708 & 0.778 & 0.704 & \underline{0.984} & 0.879 & 0.976 & \cellcolor{blue!10}0.971 & \cellcolor{blue!10}\textbf{0.983}\\
                                         & P-AUC & 0.818 & 0.882 & 0.879 & 0.957 & 0.943 & \underline{0.977} & \cellcolor{blue!10}0.973 & \cellcolor{blue!10}\textbf{0.982}\\
                                         & PRO   & 0.901 & 0.912 & 0.879 & 0.915 & 0.861 & 0.870 & \cellcolor{blue!10}\underline{0.921} & \cellcolor{blue!10}\textbf{0.939}\\
                                         \cmidrule(r){2-10}
                                         & Avg.  & 0.809 & 0.857 & 0.821 & 0.952 & 0.894 & 0.941 & \cellcolor{blue!10}\underline{0.955} & \cellcolor{blue!10}\textbf{0.968}\\
                                         \midrule
    \bottomrule
  \end{tabular}
}
\end{table*}

\begin{figure*}[!t]
\centering
\includegraphics[width=0.95\linewidth]{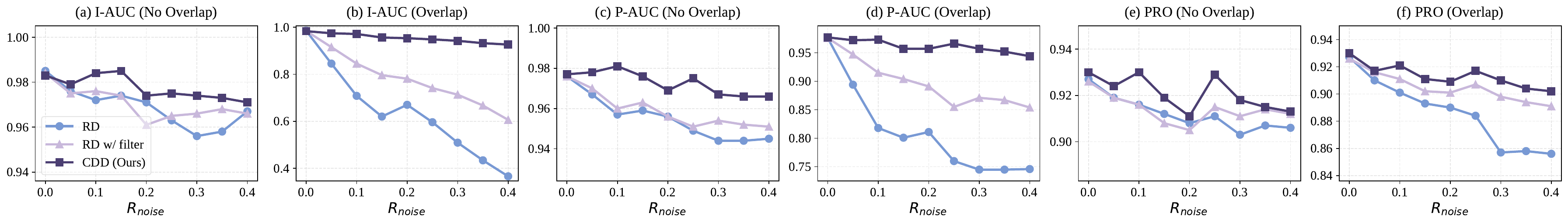}
\caption{ Comparison of anomaly detection and localization performance with baseline RD~\cite{deng2022anomaly} and RD improved by an anomaly filter (RD w/ filter) under different \( R_{\text{noise}}\) on MVTec AD~\cite{bergmann2019mvtec} in \textit{No Overlap} and \textit{Overlap} settings.}
\label{fig_r_noise}
\end{figure*}

\begin{table*}
  \caption{Anomaly detection and localization results in \textit{No Overlap}  and \textit{Overlap} settings on VisA-noise-0.05~\cite{zou2022spot} with the best in bold
  and the second best underlined.
  For fair comparison, all baseline results are reproduced in the same experimental environment. 
  }
  \label{tab_all_results:visa}
  \centering
  \resizebox{0.9\linewidth}{!}{
  \begin{tabular}{c|c|ccc|ccccc}
    \toprule
   \multicolumn{2}{c|}{Type} & \multicolumn{3}{c|}{Unsupervised} &  \multicolumn{5}{c}{Fully Unsupervised} \\
    \midrule
    Setting & Metric  & RD~\cite{deng2022anomaly}  & RD++~\cite{tien2023revisiting}  & Dinomaly~\cite{guo2025dinomaly} & SoftPatch~\cite{jiang2022softpatch} & InReaCh~\cite{mcintosh2023inter} & FUN-AD~\cite{im2025fun}  & \cellcolor{blue!10}CDD$^*$ (Ours) & \cellcolor{blue!10}CDD$^\dagger$ (Ours)\\
    \midrule
    \multirow{4}{*}{\textit{No Overlap}} & I-AUC & 0.945 & 0.912 & 0.973 & 0.927 & 0.827 & 0.924 & \cellcolor{blue!10}0.954 & \cellcolor{blue!10}\textbf{0.981}\\
                                         & P-AUC & 0.979 & 0.979 & \textbf{0.988} & 0.985 & 0.974 & 0.984 & \cellcolor{blue!10}0.982 & \cellcolor{blue!10}\underline{0.987}\\
                                         & PRO   & 0.897 & 0.898 & \textbf{0.944} & 0.904 & 0.793 & 0.797 & \cellcolor{blue!10}0.911 & \cellcolor{blue!10}\underline{0.943}\\
                                         \cmidrule(r){2-10}
                                         & Avg.  & 0.940 & 0.930 & \underline{0.968} & 0.939 & 0.865 & 0.902 & \cellcolor{blue!10}0.949 & \cellcolor{blue!10}\textbf{0.970}\\
                                    \midrule
    \multirow{4}{*}{\textit{Overlap}}    & I-AUC & 0.656 & 0.798 & 0.721 & 0.924 & 0.725 & 0.934 & \cellcolor{blue!10}\underline{0.936} & \cellcolor{blue!10}\textbf{0.976}\\
                                         & P-AUC & 0.909 & 0.960 & 0.972 & 0.954 & 0.914 & 0.982 & \cellcolor{blue!10}\underline{0.977} & \cellcolor{blue!10}\textbf{0.986}\\
                                         & PRO   & 0.892 & 0.894 & \underline{0.928} & 0.883 & 0.721 & 0.793 & \cellcolor{blue!10}0.911 & \cellcolor{blue!10}\textbf{0.945}\\  
                                         \cmidrule(r){2-10}
                                         & Avg.  & 0.819 & 0.884 & 0.874 & 0.920 & 0.787 & 0.903 & \cellcolor{blue!10}\underline{0.941} & \cellcolor{blue!10}\textbf{0.969}\\
                                         \midrule
    \bottomrule
  \end{tabular}
  }
\end{table*}

\subsection{Anomaly Detection under FUAD setting}

\subsubsection{Quantitative Results on MVTec AD}
We evaluate our proposed Cross-\textcolor{hl}{Division} Distillation (CDD) on the MVTec-AD dataset with \( R_{\text{noise}} = 0.1 \), denoted as MVTec-AD-noise-0.1. 
CDD is compared with unsupervised KD-based UAD methods including RD~\cite{deng2022anomaly}, RD++~\cite{tien2023revisiting}, and Dinomaly~\cite{guo2025dinomaly}, along with FUAD methods, such as SoftPatch~\cite{jiang2022softpatch}, InReaCh~\cite{mcintosh2023inter}, and FUN-AD~\cite{im2025fun}. Table~\ref{tab_all_results:mvtec} presents the anomaly detection and localization results under \textit{No Overlap} and \textit{Overlap} settings, respectively, where each method reports I-AUC, P-AUC, and PRO metrics, all reproduced through 200 epochs of model training under a unified dataset split.  

In the \textit{No Overlap} setting, our RD-based implementation CDD$^*$ already matches SoftPatch's I-AUC while achieving a P-AUC of 0.981 and PRO of 0.930, surpassing previous methods in pixel-level localization. Furthermore, by incorporating our training strategy into Dinomaly, CDD$^\dagger$ sets a new SOTA with an I-AUC of 0.992, P-AUC of 0.986, and PRO of 0.941.

In the \textit{Overlap} setting, despite the performance of baselines including RD and Dinomaly dropping sharply to around 0.7 in I-AUC due to noise interference, our methods retain exceptional robustness. Even CDD$^*$, based on RD, achieves a P-AUC of 0.973 and PRO of 0.921, which is comparable to previous SOTA methods and outperforms SoftPatch in localization. Moreover, CDD$^\dagger$ significantly revitalizes the baseline Dinomaly, improving I-AUC by 27.9\% (from 0.704 to 0.983) and substantially elevating the SOTA performance across all metrics.
\textcolor{hl}{
Since CDD applies no setting-specific operation for the \textit{Overlap} setting, these improvements indicate that the cross-division training strategy implicitly suppresses the memorization of anomalous samples encountered during training, allowing them to remain detectable when they reappear at test time.
}

\textbf{Results under Different Noise Ratios.}
To further validate the robustness of our method, we compare CDD against  RD and an RD variant with the filtering strategy (denoted as RD w/ Filter) 
across a wide range of anomaly ratios \( R_{\text{noise}} \)  from 0 to 0.4 on MVTec AD in \textit{No Overlap} and \textit{Overlap} setting, as in Fig.~\ref{fig_r_noise}.
The results demonstrate a clear trend: as \( R_{\text{noise}} \)  increases, the performance of both the baseline RD and RD w/ filter degrades significantly. In contrast, our CDD  maintains consistent and superior performance across all three evaluation metrics (I-AUC, P-AUC, PRO) with small fluctuations. This demonstrates that our ``\textit{isolating to harness}" strategy effectively controls anomaly interference without discarding any training data, achieving robust generalization under severe contamination.

\begin{figure*}[!t]
\centering
\includegraphics[width=0.95\linewidth]{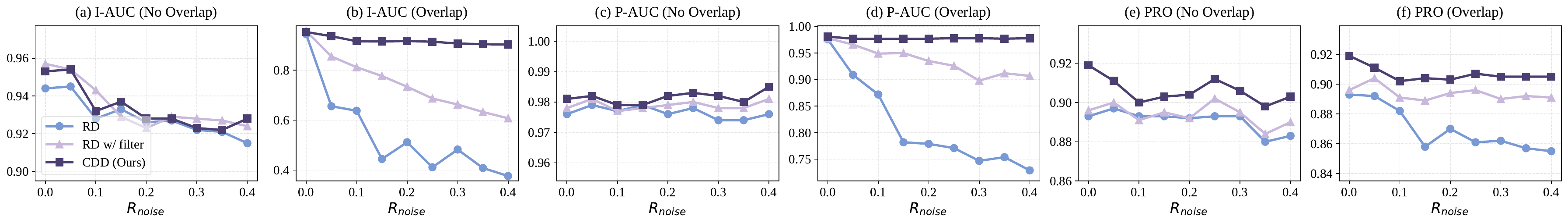}
\caption{ Comparison of anomaly detection and localization performance with baseline RD~\cite{deng2022anomaly} and RD improved by an anomaly filter (RD w/ filter) under different \( R_{\text{noise}}\) on VisA~\cite{zou2022spot} in \textit{No Overlap} and \textit{Overlap} settings.}
\label{fig_r_noise_visa}
\end{figure*}

\begin{table*}
  \caption{Anomaly detection and localization results in \textit{No Overlap}  and \textit{Overlap} settings on MVTec 3D-AD-noise-0.1~\cite{bergmann2021mvtec} with the best in bold
  and the second best underlined.
  For fair comparison, all baseline results are reproduced in the same experimental environment. 
  }
  \label{tab_all_results:mvtec_3d}
  \centering
  \resizebox{0.9\linewidth}{!}{
  \begin{tabular}{c|c|ccc|ccccc}
    \toprule
   \multicolumn{2}{c|}{Type} & \multicolumn{3}{c|}{Unsupervised} &  \multicolumn{5}{c}{Fully Unsupervised} \\
    \midrule
    Setting & Metric  & RD~\cite{deng2022anomaly}  & RD++~\cite{tien2023revisiting}  & Dinomaly~\cite{guo2025dinomaly} & SoftPatch~\cite{jiang2022softpatch} & InReaCh~\cite{mcintosh2023inter} & FUN-AD~\cite{im2025fun}  & \cellcolor{blue!10}CDD$^*$ (Ours) & \cellcolor{blue!10}CDD$^\dagger$ (Ours)\\
    \midrule
    \multirow{4}{*}{\textit{No Overlap}} & I-AUC & 0.837 & 0.830 & \textbf{0.910} & 0.800 & 0.598 & 0.801 & \cellcolor{blue!10}0.847 & \cellcolor{blue!10}\textbf{0.910}\\
                                         & P-AUC & 0.985 & 0.986 & \textbf{0.991} & 0.982 & 0.964 & 0.970 & \cellcolor{blue!10}0.986 & \cellcolor{blue!10}\textbf{0.991}\\
                                         & PRO   & 0.947 & 0.953 & \underline{0.964} & 0.941 & 0.878 & 0.833 & \cellcolor{blue!10}0.952 & \cellcolor{blue!10}\textbf{0.969}\\
                                         \cmidrule(r){2-10}
                                         & Avg.  & 0.923 & 0.923 & \underline{0.955} & 0.908 & 0.813 & 0.868 & \cellcolor{blue!10}0.928 & \cellcolor{blue!10}\textbf{0.957}\\
                                    \midrule
    \multirow{4}{*}{\textit{Overlap}}    & I-AUC & 0.600 & 0.645 & 0.652 & 0.757 & 0.553 & 0.804 & \cellcolor{blue!10}\underline{0.818} & \cellcolor{blue!10}\textbf{0.872}\\
                                         & P-AUC & 0.856 & 0.947 & 0.895 & 0.933 & 0.941 & 0.970 & \cellcolor{blue!10}\underline{0.976} & \cellcolor{blue!10}\textbf{0.986}\\
                                         & PRO   & 0.935 & \underline{0.951} & 0.905 & 0.872 & 0.866 & 0.836 & \cellcolor{blue!10}0.948 & \cellcolor{blue!10}\textbf{0.966}\\ 
                                         \cmidrule(r){2-10}
                                         & Avg.  & 0.797 & 0.848 & 0.817 & 0.854 & 0.787 & 0.870 & \cellcolor{blue!10}\underline{0.914} & \cellcolor{blue!10}\textbf{0.941}\\
                                         \midrule
    \bottomrule
  \end{tabular}
  }
\end{table*}

\begin{figure*}
  \centering
  \includegraphics[width=0.9\linewidth]{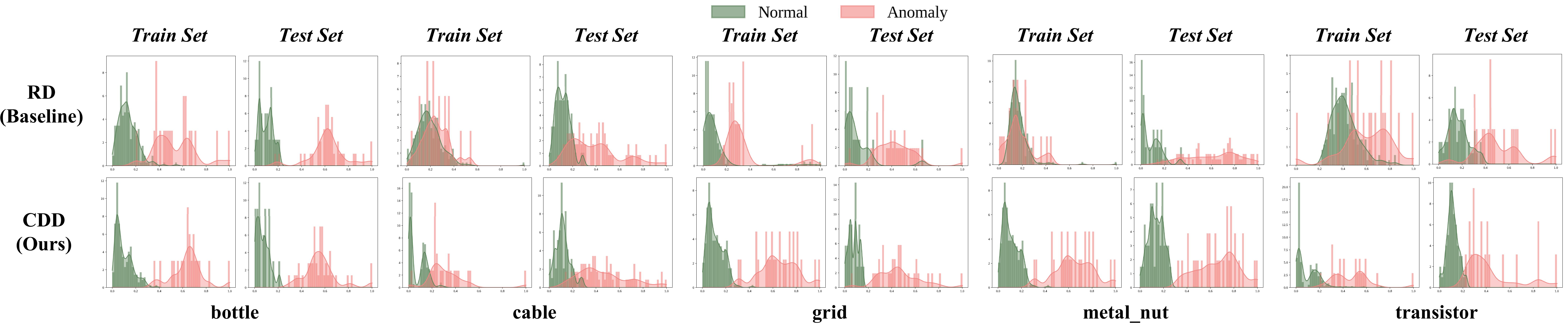}
  \caption{Comparison of histograms of anomaly scores obtained by RD~\cite{deng2022anomaly} and CDD on MVTec AD-noise-0.1~\cite{bergmann2019mvtec}.
  }
  \label{fig_hist}
\end{figure*}

\subsubsection{Quantitative Results on VisA}
For the VisA dataset, we set \( R_{\text{noise}} = 0.05 \) (VisA-noise-0.05) based on the ratio of normal to anomalous samples in the original dataset and conduct relevant experiments as in Table~\ref{tab_all_results:visa}. The compared methods include unsupervised and fully unsupervised AD methods.

In the \textit{No Overlap} setting, 
CDD$^*$, based on RD, consistently improves upon the vanilla RD across all metrics.
For the Dinomaly-based implementation, CDD$^\dagger$ further boosts the image-level detection capability, improving the I-AUC from 0.973 to 0.981, while maintaining highly competitive pixel-level localization performance comparable to the pure baseline.
In the \textit{Overlap} setting, the advantage of our approach becomes even more critical. While competing FUAD methods like SoftPatch and FUN-AD maintain reasonable I-AUC scores, their localization precision drops significantly.
In contrast, CDD$^*$ and CDD$^\dagger$ maintain exceptional localization robustness with PRO scores of 0.911 and 0.944, respectively. Furthermore, CDD$^*$ and CDD$^\dagger$  outperform the vanilla RD and Dinomaly by a substantial margin in I-AUC (0.656 $\to$ 0.936 and 0.721 $\to$ 0.975), 
demonstrating that our cross-division training strategy effectively enhances the baseline's resilience to anomaly interference.

\textbf{Results under Different Noise Ratios.}
We compare the robustness of CDD against RD and RD w/ filter on VisA under varying noise ratios \( R_{\text{noise}} \). As illustrated in Fig.~\ref{fig_r_noise_visa}, while the performance of RD and RD w/ filter tends to fluctuate or degrade as the noise ratio \( R_{\text{noise}} \) increases, CDD demonstrates superior stability, especially in the \textit{Overlap} setting. In most metrics, particularly under high noise levels, CDD consistently outperforms both RD and RD w/ filter, further validating the generalizability of our purification strategy on highly contaminated datasets.

\subsubsection{Quantitative Results on MVTec 3D-AD}

To verify the generalization of our framework on diverse modalities, we conduct experiments on the MVTec 3D-AD (RGB) dataset with \( R_{\text{noise}} = 0.1 \), as in Table~\ref{tab_all_results:mvtec_3d}.
In the \textit{No Overlap} setting, CDD$^\dagger$ achieves the highest performance with an PRO of 0.969, demonstrating that our strategy further refines localization boundaries.
In the \textit{Overlap} setting, baselines suffer severe degradation. The I-AUC of RD and Dinomaly collapses to 0.600 and 0.652, respectively. 
However, our framework successfully revitalizes these baselines: CDD$^*$ boosts the I-AUC of RD by over 21.8\%, and CDD$^\dagger$ improves the baseline Dinomaly by 22.0\%. 
Moreover, in terms of pixel-level localization, both variants achieve state-of-the-art P-AUC scores, significantly surpassing SoftPatch and FUN-AD. 
This confirms that CDD is capable of learning robust normal representations even in challenging industrial scenarios.

\begin{figure*}
  \centering
  \includegraphics[width=0.9\linewidth]{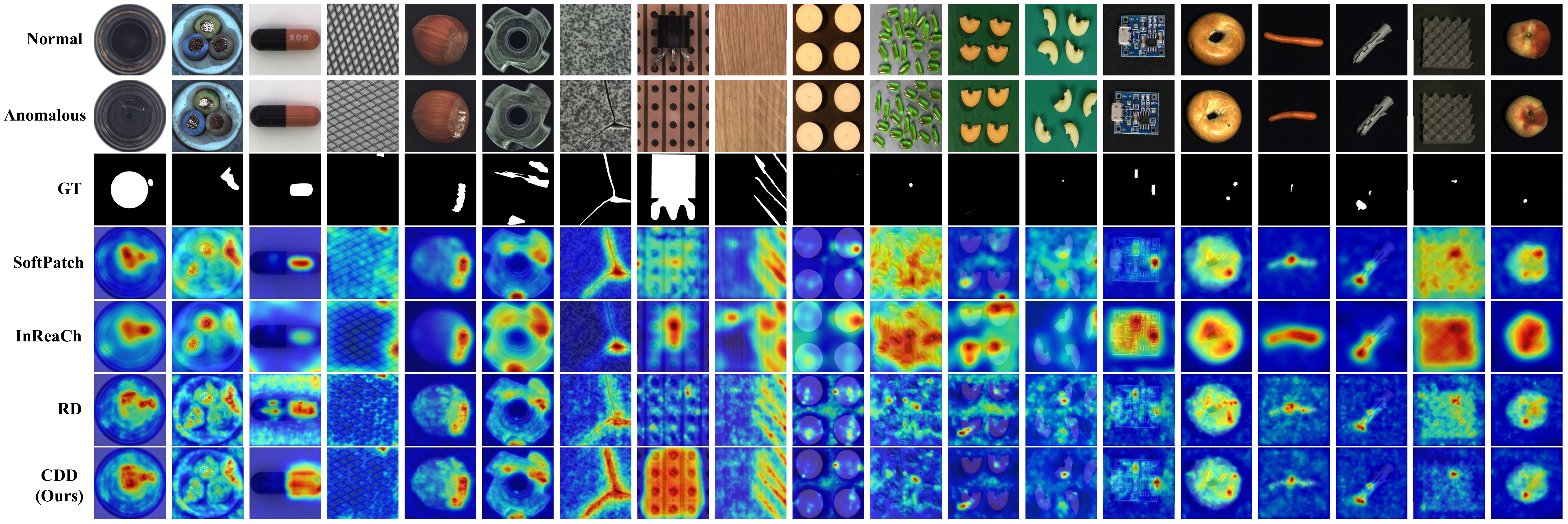}
  \caption{
  Qualitative comparison of anomaly maps generated by SoftPatch~\cite{jiang2022softpatch}, InReaCh~\cite{mcintosh2023inter}, RD~\cite{deng2022anomaly} and our CDD
  on MVTec AD-noise-0.1~\cite{bergmann2019mvtec}, VisA-noise-0.05~\cite{zou2022spot}, and MVTec 3D-AD-noise-0.1~\cite{bergmann2021mvtec} in the \textit{No Overlap} setting.
  }
  \label{fig_compare}
\end{figure*}

\begin{figure}
  \centering
  \includegraphics[width=\linewidth]{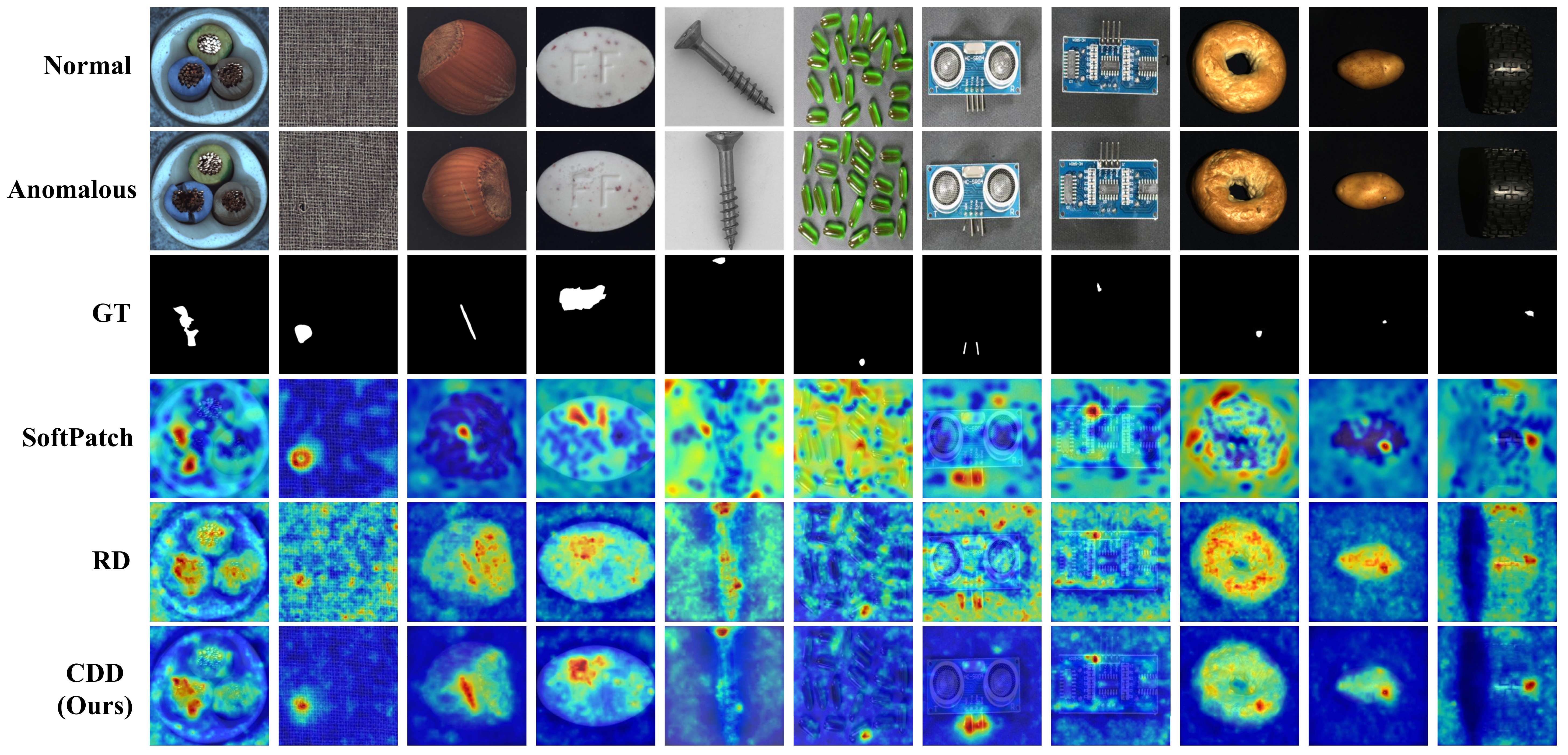}
  \caption{
  Qualitative comparison of anomaly maps generated by SoftPatch~\cite{jiang2022softpatch}, InReaCh~\cite{mcintosh2023inter}, RD~\cite{deng2022anomaly} and our CDD
  on MVTec AD-noise-0.1~\cite{bergmann2019mvtec}, VisA-noise-0.05~\cite{zou2022spot}, and MVTec 3D-AD-noise-0.1~\cite{bergmann2021mvtec} in the \textit{Overlap} setting. The displayed test samples also appear in the training set as unlabeled noise. 
  }
  \label{fig_compare_overlap}
\end{figure}

\begin{table}[t] \color{hl}
  \caption{\textcolor{hl}{Training time and peak GPU memory for complete 200-epoch training runs on the \textit{bottle} category of MVTec AD-noise-0.1~\cite{bergmann2019mvtec}. All settings use the same hardware, batch size, and input resolution.}}
  \label{tab_training_cost}
  \centering
  \resizebox{0.75\columnwidth}{!}{
  \begin{tabular}{c|c|cc}
    \toprule
    Method & \(K\) Schedule
    & Time (min) 
    & Peak GPU Mem. (MiB)  \\
    \midrule
    RD~\cite{deng2022anomaly}  & -- & 8.86  & 5192  \\
    CDD & \(2\)               & 37.97 & 8318   \\
    CDD & \(3\)               & 49.39  & 9920   \\
    CDD & \(4\)               & 63.30  & 11488 \\
   \rowcolor{blue!10} CDD & \(\{2,3,3,2\}\)     & 44.68  &9916  \\
    \bottomrule
  \end{tabular}
  }
\end{table}

\begin{figure}
  \centering
  \includegraphics[width=0.8\linewidth]{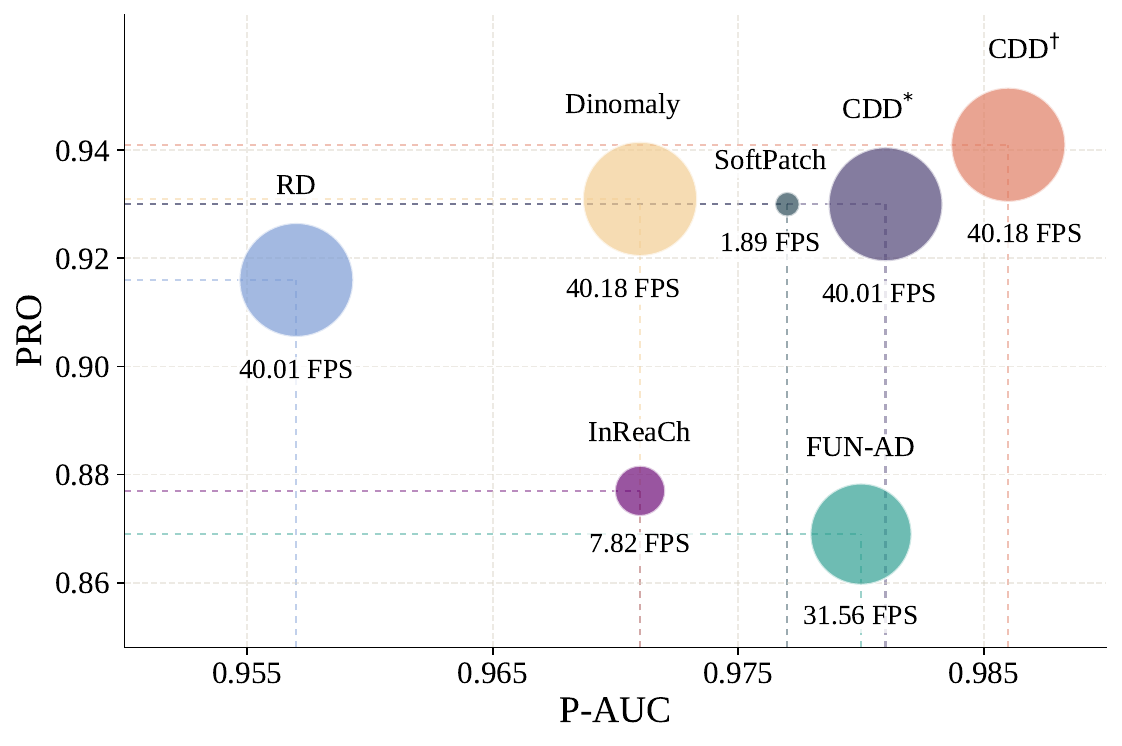}
  \caption{Comparisons of anomaly localization performance and inference speed. The results are reported on MVTec AD-noise-0.1~\cite{bergmann2019mvtec}. We visualize the Frames Per Second (FPS) as the bubble size. 
  }
  \label{fig_compare_fps}
\end{figure}

\subsubsection{Qualitative Comparisons}
We perform additional visualization experiments to compare our proposed CDD with the baseline RD. First, we obtain anomaly scores on both the training and test sets of MVTec-AD-noise-0.1 using the trained RD and CDD, generating histograms of anomaly scores for all the samples as depicted in Fig.~\ref{fig_hist}. RD proves effective in the FUAD setting, yet it inadvertently learns certain anomaly patterns from the training set, weakening its ability to accurately detect anomalies. Notably, our CDD overcomes this limitation, markedly improving anomaly detection ability even on the training set.

Fig.~\ref{fig_compare} further compares anomaly maps generated by SoftPatch~\cite{jiang2022softpatch}, InReaCh~\cite{mcintosh2023inter}, RD~\cite{deng2022anomaly}, and our CDD. The visualization shows that CDD generates anomaly maps 
with fewer false negatives in anomalous regions and reduced false positives in normal areas.
In contrast, the baseline RD tends to miss some subtle anomalies. These results intuitively demonstrate CDD's superior anomaly sensitivity and precision in anomaly localization.

In FUAD task, it is challenging for a trained model to localize anomalies for the anomalous samples seen during the training phase, which lead to the introduction of the Overlap setting. Therefore, we also conduct visualizations by generating anomaly maps for the anomalous samples selected from the training set. Fig.~\ref{fig_compare_overlap} compares anomaly maps from SoftPatch~\cite{jiang2022softpatch}, the baseline RD~\cite{deng2022anomaly} and our CDD. The results show that our method, through cross-division training, effectively mitigates the noise interference from training set and achieves accurate anomaly localization even on seen anomalous samples.

\begingroup
\color{hl}
\subsubsection{Computational Efficiency}

To provide a complete analysis of computational efficiency, Table~\ref{tab_training_cost} reports the training time and peak GPU memory for complete 200-epoch runs on the \textit{bottle} category of MVTec AD-noise-0.1~\cite{bergmann2019mvtec} under identical settings. GPU memory is sampled at 2-second intervals. Compared with RD~\cite{deng2022anomaly}, CDD requires additional training time and memory, both of which increase with \(K\) due to the division-specific students and cross-division feature generation. However, this overhead is confined to offline training, since all division-specific students are discarded after training and only the global student is retained for inference.
\endgroup

Considering that high inference speed is critical for industrial deployment, we evaluate the inference speed of our method compared to existing FUAD methods. As in Fig.~\ref{fig_compare_fps}, CDD achieves a favorable trade-off between accuracy and efficiency. It significantly outperforms previous FUAD methods in terms of Frames Per Second (FPS) while achieving state-of-the-art detection capabilities, demonstrating its potential for real-time anomaly detection applications.

\begin{table*}[t]
\centering

\begin{minipage}[t]{0.67\textwidth}
\vspace{0pt}
\centering
  \caption{
  \textcolor{hl}{
  Ablation study of module effectiveness on MVTec AD-noise-0.1~\cite{bergmann2019mvtec} with the dynamic division schedule \(K=\{2,3,3,2\}\) in \textit{No Overlap} and \textit{Overlap} settings.}}
  \label{tab_ablation}
  \centering
  \resizebox{0.98\linewidth}{!}{
  \begin{tabular}{ccc|ccc|ccc|ccc|c}
    \toprule
    \multicolumn{3}{c|}{DST} & \multicolumn{3}{c|}{CDKA}
    & \multicolumn{3}{c|}{\textit{No Overlap}}
    & \multicolumn{3}{c|}{\textit{Overlap}}
    & \multirow{2}{*}{\textit{Average}} \\
    \cmidrule(r){1-3}
    \cmidrule(r){4-6}
    \cmidrule(r){7-9}
    \cmidrule(r){10-12}
    \textit{D.C.} & \textit{Conf.G.} & \textit{Reg.}
    & \textit{P.N.} & \textit{F.P.} & \textit{Conf.D.}
    & I-AUC & P-AUC & PRO
    & I-AUC & P-AUC & PRO & \\
    \midrule
    - & - & - & - & - & -
    & 0.9721 & 0.9566 & 0.9156
    & 0.7084 & 0.8184 & 0.9007
    & 0.8786 \\
    \checkmark & - & - & \checkmark & - & -
    & 0.9671 & 0.9727 & 0.9215
    & 0.9632 & 0.9674 & 0.9190
    & 0.9518 \\
    \checkmark & \checkmark & - & \checkmark & - & -
    & 0.9728 & 0.9752 & 0.9167
    & 0.9711 & 0.9703 & 0.9122
    & 0.9531 \\
    \checkmark & \checkmark & - & \checkmark & - & \checkmark
    & 0.9793 & 0.9785 & 0.9197
    & 0.9655 & 0.9689 & 0.9124
    & 0.9540 \\
    \checkmark & \checkmark & - & \checkmark & \checkmark & \checkmark
    & 0.9816 & 0.9807 & 0.9279
    & 0.9701 & 0.9721 & 0.9199
    & 0.9587 \\
    \rowcolor{blue!10}\checkmark & \checkmark & \checkmark & \checkmark & \checkmark & \checkmark
    & 0.9840 & 0.9812 & 0.9297
    & 0.9713 & 0.9726 & 0.9207
    & 0.9599 \\
    \bottomrule
  \end{tabular}
  }
\end{minipage}
\hfill
\begin{minipage}[t]{0.32\textwidth}
\vspace{0pt}
\centering
   \caption{Ablation study of division number \(K\) on MVTec AD-noise-0.1~\cite{bergmann2019mvtec}.
  }
  \label{tab_K}
  \centering
  \resizebox{0.98\linewidth}{!}{
  \begin{tabular}{c|cccc}
    \toprule
 \(K\) & I-AUC& P-AUC& PRO & Average\\
    \midrule
    2 & 0.9836 & 0.9818 & 0.9287 & 0.9647\\
    3 & 0.9821 & 0.9793 & 0.9252 & 0.9622 \\
    4 & 0.9791 & 0.9811 & 0.9271 & 0.9624 \\
    \color{hl} \{2,3\} & \color{hl} 0.9808 & \color{hl} 0.9813 & \color{hl} 0.9293 & \color{hl} 0.9638\\
    \color{hl} \{3,2\} & \color{hl} 0.9579 & \color{hl} 0.9633 & \color{hl} 0.9197 & \color{hl} 0.9470\\
 \rowcolor{blue!10}   \{2,3,3,2\} & 0.9840 & 0.9812 & 0.9297 & 0.9650\\
    \{2,3,4,3,2\} & 0.9837& 0.9806 & 0.9260 & 0.9634 \\
    \bottomrule
  \end{tabular}
  }
\end{minipage}

\end{table*}

\begin{table}[t]
  \caption{Ablation study of pseudo-normal feature selection strategies on MVTec AD-noise-0.1~\cite{bergmann2019mvtec} with \(K=3\).}
  \label{tab_select}
  \centering
  \resizebox{0.8\columnwidth}{!}{
  \begin{tabular}{c|l|cccc}
    \toprule
   \multicolumn{6}{c}{\(K=3\)} \\
    \midrule
  \multicolumn{2}{c|}{Selection Strategy} & I-AUC& P-AUC& PRO & Average\\
    \midrule
   \multicolumn{2}{c|}{\textit{All}} & 0.9753& 0.9747 &0.9251 & 0.9584 \\
    \midrule
   \multirow{2}{*}{One} & \textit{Next} & 0.9510 &0.9692 &0.9142& 0.9448\\
   & \cellcolor{blue!10}\textit{Global-Guided} & \cellcolor{blue!10}0.9821 & \cellcolor{blue!10}0.9793 & \cellcolor{blue!10}0.9252 & \cellcolor{blue!10}0.9622\\
    \bottomrule
  \end{tabular}
  }
\end{table}

\subsection{Ablation Analysis}

\textcolor{hl}{
All ablation studies are conducted using the RD-based implementation, denoted as CDD\(^*\). Unless otherwise specified, the experiments are performed on MVTec AD-noise-0.1~\cite{bergmann2019mvtec} in the \textit{No Overlap} setting.
}

\subsubsection{Effectiveness of Proposed Designs}
\textcolor{hl}{
We conduct stepwise ablation experiments on MVTec AD-noise-0.1~\cite{bergmann2019mvtec} in both the \textit{No Overlap} and \textit{Overlap} settings, as reported in Table~\ref{tab_ablation}. All CDD variants adopt the final dynamic division schedule \(K=\{2,3,3,2\}\). The configuration without any proposed component reverts to the baseline RD. We then incrementally integrate the proposed modules to evaluate their individual contributions.
}

\begin{table}[t]
\color{hl}
\centering
\caption{
\textcolor{hl}{
Ablation study of the high-confidence ratio schedule on MVTec
AD-noise-0.1~\cite{bergmann2019mvtec}.
}}
\label{tab:r_schedule}
\resizebox{0.7\columnwidth}{!}{
\begin{tabular}{cc|cccc}
\toprule
$r_{\max}$ & $e_s$ & I-AUC & P-AUC & PRO & Average \\
\midrule
0.3 & 0   & 0.9730 & 0.9755 & 0.9213 & 0.9566 \\
0.4 & 0   & 0.9782 & 0.9791 & 0.9223 & 0.9598 \\
\rowcolor{blue!10} 0.5 & 0   & 0.9840 & 0.9812 & 0.9297
          & 0.9650 \\
0.6 & 0   & 0.9825 & 0.9801 & 0.9283 & 0.9636 \\
0.7 & 0   & 0.9813 & 0.9814 & 0.9298
          & 0.9642 \\
\midrule
0.5 & 20  & 0.9725 & 0.9789 & 0.9247 & 0.9587 \\
0.5 & 50  & 0.9639 & 0.9761 & 0.9199 & 0.9533 \\
0.5 & 100 & 0.9649 & 0.9774 & 0.9265 & 0.9563 \\
\bottomrule
\end{tabular}
}
\end{table}

For Division-Specific Training (DST), we evaluate: (1) \textit{D.C.}: A naive data partitioning strategy that evenly splits the training set into data divisions. 
(2) \textit{Conf.G.}: Our proposed Confidence-Guided Division Construction, which actively lowers the anomaly ratio in each division.
(3) \textit{Reg.}: The regularization term in the division-specific student loss, which encourages consistency with the global student from the previous epoch.

For Cross-Division Knowledge Aggregation (CDKA), we assess:
(1) \textit{P.N.}: The basic cross-division distillation, generating pseudo-normal features across divisions for feature distillation.
(2) \textit{F.P.}: Applying feature perturbation to the global student’s input during training.
(3) \textit{Conf.D.}: Confident Distillation providing direct teacher supervision on high-confidence normal samples.

\textcolor{hl}{
The results in Table~\ref{tab_ablation} demonstrate the complementary effects of the proposed designs. Introducing \textit{D.C.} and \textit{P.N.} substantially improves performance in the \textit{Overlap} setting, increasing I-AUC and P-AUC from \(0.7084\) and \(0.8184\) to \(0.9632\) and \(0.9674\), respectively. This improvement directly verifies that Cross-Division Pseudo-Normal Feature Distillation effectively suppresses the memorization of training anomalies. Although individual metrics may fluctuate when intermediate components are introduced, the overall \textit{Average} increases progressively from \(0.8786\) for the RD baseline to \(0.9599\) for the complete model. In particular, the full configuration achieves the best results across all metrics in both settings, validating the complementary contributions of the proposed components.
}

\begin{figure*}
  \centering
  \includegraphics[width=0.9\linewidth]{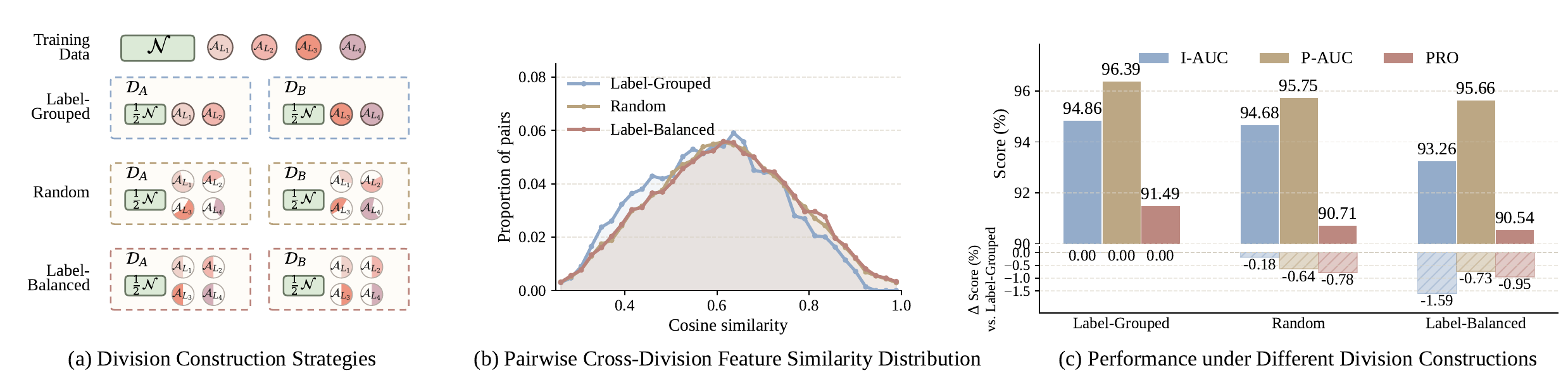}
\caption{
\textcolor{hl}{
Controlled analysis of division construction on MVTec AD-noise-0.4~\cite{bergmann2019mvtec} (excluding \textit{toothbrush}, which lacks semantic anomaly labels).
    (a) \textit{Label-Grouped}, \textit{Random}, and \textit{Label-Balanced} constructions.
    (b) Pairwise cosine-similarity distributions of cross-division teacher features extracted from anomalous regions.
    (c) I-AUC, P-AUC, and PRO under the three constructions.
    Ground-truth anomaly labels are used only to construct the diagnostic settings.
    }
}
  \label{fig:anomaly_similarity}
\end{figure*}

\subsubsection{Number of Divisions}

To investigate the impact of the number of divisions \( K \), we conduct an ablation study on MVTec-AD-noise-0.1~\cite{bergmann2019mvtec}, with performance results in Table~\ref{tab_K}. 
For a dynamic schedule, the listed \(K\) values are applied sequentially over equal-length training intervals.
Theoretically, increasing \( K \) allows finer control over the number of anomaly patterns within each data division.
However, a larger \( K \) reduces the sample size per division, weakening the division-specific students' ability to generate normal features.
Thus, a balance between anomaly suppression and normal feature modeling is needed. Experimental results in Table~\ref{tab_K} support this hypothesis: as \( K \) increases from 2 to 4, the average performance drops from 0.9647 to 0.9624, indicating that \( K = 2 \) achieves a good balance.

As training progresses, the student can gradually generate normal features. In this case, appropriately increasing \(K\) better isolates anomalies. In the later stages, as the global student serving as a regularizer in division-specific training learns to generate normal features even in anomaly regions, finer division partitioning becomes less critical, allowing \( K \) to be reduced.
\textcolor{hl}{
To validate this scheduling rationale, we compare increasing, decreasing, and increase-then-decrease schedules in Table~\ref{tab_K}. The clear advantage of \(\{2,3\}\) over \(\{3,2\}\) indicates that starting with a small \(K\) is important for stable early training. Increasing \(K\) further to \(4\), as in \(\{2,3,4,3,2\}\), provides no additional benefit. In contrast, \(\{2,3,3,2\}\) achieves the highest PRO of 0.9297, supporting a moderate increase in \(K\) during intermediate training followed by a reduction in the final stage.
}

\textcolor{hl}{
Taken together, these results suggest that \(K\) should balance anomaly-pattern separation against stable division-specific learning. Small values of \(K\) retain sufficient samples in each division, while a moderate increase during intermediate training enhances cross-division diversity.
}
Therefore, our final design adopts \(K=\{2,3,3,2\}\) across epochs.

\subsubsection{Selection of Pseudo-Normal Features}

We conduct an ablation study on pseudo-normal feature selection strategies, all performed with \(K=3\), with results presented in Table~\ref{tab_select}. One strategy, labeled \textit{All}, uses pseudo-normal features generated by division-specific students from all other data divisions for distillation. Alternatively, we select features from only one division, either via our Global-Guided Pseudo-Normal Feature Selection (denoted as \textit{Global-Guided}) or by choosing the next division's feature (denoted as \textit{Next}, akin to random selection).
\textcolor{hl}{
The comparison is conducted with \(K=3\) because it is the only setting in our final division schedule \(\{2,3,3,2\}\) that requires explicit selection among multiple out-of-division candidates. When \(K=2\), each sample has only one such candidate, and the three selection strategies consequently degenerate into the same operation. 
}
The results show that our \textit{Global-Guided} strategy markedly achieves the best performance, demonstrating that the Consensus-driven strategy significantly enhances the quality of cross-division distillation.

\begingroup
\color{hl}

\subsubsection{Analysis of the High-Confidence Ratio Schedule}
\label{sec:r_schedule}

The high-confidence ratio schedule \(r(e)\) is jointly determined by the maximum ratio \(r_{\max}\) and the activation epoch \(e_s\). We analyze these two factors on MVTec AD-noise-0.1~\cite{bergmann2019mvtec} by varying one parameter while fixing the other to its default value.
As shown in Table~\ref{tab:r_schedule}, the performance remains relatively stable when \(r_{\max}\) varies from 0.3 to 0.7. Increasing \(r_{\max}\) from 0.3 to 0.5 consistently improves all metrics, indicating that sharing more high-confidence samples provides stronger normal support for the division-specific students. Further increasing \(r_{\max}\) yields only marginal changes: \(r_{\max}=0.7\) achieves slightly higher P-AUC and PRO, whereas the default \(r_{\max}=0.5\) obtains the best I-AUC and average performance. We therefore adopt \(r_{\max}=0.5\) as the default setting.

We further delay confidence guidance by setting \(r(e)=0\) before epoch \(e_s\). All delayed settings perform worse than \(e_s=0\), indicating that the gradual increase of \(r(e)\) already serves as an effective warm-up. Further postponing confidence guidance reduces its effective training duration and consequently degrades performance.

\endgroup

\begingroup
\color{hl}

\subsection{Controlled Analysis of Cross-Division Anomaly Similarity}
\label{sec:anomaly_similarity}

To examine the potential effect of cross-division anomaly similarity on CDD, we conduct a controlled experiment on MVTec AD-noise-0.4~\cite{bergmann2019mvtec}. 
In particular, we analyze how assigning semantically related anomalous samples to the same or different divisions affects the anomaly-suppression effect of cross-division training.
We use a simplified two-stage protocol to exclude the influence of other CDD components: two students are independently trained on their respective divisions for 20 epochs, followed by 20 epochs of global-student training using their cross-division outputs.

As 
in Fig.~\ref{fig:anomaly_similarity} (a), we compare three division construction strategies. \textit{Label-Grouped} places anomalies sharing the same anomaly label in the same division whenever possible, \textit{Random} randomly assigns anomalous samples, and \textit{Label-Balanced} evenly distributes each anomaly label across the two divisions. Normal samples are equally divided, and ground-truth anomaly labels are used only to construct these settings.

Fig.~\ref{fig:anomaly_similarity} (b) shows the pairwise cosine-similarity distributions of teacher features extracted from ground-truth anomalous regions in different divisions. The distributions shift only slightly from \textit{Label-Grouped} to \textit{Random} and \textit{Label-Balanced}, indicating instance-level diversity even among anomalies sharing the same semantic label. The performance differences in Fig.~\ref{fig:anomaly_similarity} (c) are also limited: \textit{Random} decreases the average performance by only 0.53\% relative to \textit{Label-Grouped}, while \textit{Label-Balanced} causes a further moderate decrease. 
Therefore, higher cross-division anomaly similarity can weaken the anomaly-suppression effect of cross-division training, but strict label-wise separation is not necessary for the cross-division mechanism to remain effective.

\endgroup

\section{Conclusion} \label{sec:conclusion}

In this paper, we propose a novel Cross-Division Distillation framework, pioneering the application of the Knowledge Distillation paradigm to the FUAD task. 
The core of our approach lies in an ``isolate to harness" strategy, implemented through two coordinated components: \textcolor{hl}{Division}-Specific Training, which divisions the data to confine anomalies and trains specialized students, and Cross-\textcolor{hl}{Division} Knowledge Aggregation, which leverages these students to collaboratively generate robust pseudo-normal features for guiding a global model.
Extensive experiments on MVTec AD, VisA, and MVTec 3D-AD demonstrate that our method significantly enhances the robustness of the foundational Reverse Distillation baseline,  proving the effectiveness of transforming anomalous data from a source of interference into a resource for learning.

\textbf{Limitations.} 
Currently, our Confidence-Guided Division Construction employs a stochastic strategy to partition low-confidence samples. By leveraging the inherent sparsity and diversity of anomaly patterns alongside iterative epoch-wise reshuffling, this design effectively isolates anomalies without complex pre-processing. While this probabilistic approach proves empirically robust and computationally efficient in our experiments, 
\textcolor{hl}{
it does not explicitly exploit feature similarity to prevent highly similar or near-duplicate anomalous samples from being assigned to different divisions. Such assignments may increase the cross-sample transferability of anomaly reconstruction and consequently weaken the anomaly-suppression effect of Cross-Division Knowledge Aggregation.
}

\textbf{Future Work.} 
In future research, we aim to investigate advanced pre-processing techniques, such as clustering, to further refine the division construction process and enhance the stability of anomaly isolation. Additionally, while our current implementation is established on the Reverse Distillation paradigm, the underlying ``isolating to harness'' philosophy is generic. We plan to extend this cross-division collaboration mechanism to broader anomaly detection paradigms to further validate its generalization capability.

\bibliographystyle{IEEEtran}
\bibliography{IEEEabrv,ref}

\newpage


\twocolumn[
\begin{@twocolumnfalse}
    \centering

    {\huge Supplementary Material for\\[0.3em]
    \textit{Isolating to Harness: Cross-Division Distillation \\ for Fully Unsupervised Anomaly Detection}\par}

    \vspace{1.5em}
\end{@twocolumnfalse}
]

\vspace{1em}

 \renewcommand\thefigure{A\arabic{figure}}
 \renewcommand\thetable{A\arabic{table}}  
 \renewcommand\theequation{A\arabic{equation}}
 \setcounter{equation}{0}
 \setcounter{table}{0}
 \setcounter{figure}{0}
\setcounter{section}{0}

\section{Detailed Analysis of the Design Rationale}
\label{sec:supp_rationale}

The two assumptions introduced in Section~III-B summarize the empirical behaviors of RD models trained on contaminated data. In this section, we provide detailed analyses of the optimization mechanisms underlying these assumptions.

\subsection{Normal-Dominant Convergence}
\label{sec:supp_a1}

\textbf{Analysis.}
This assumption is explained from the Empirical Risk Minimization (ERM) perspective.
Let $\Omega_{\mathcal{N}}$ and $\Omega_{\mathcal{A}}$ denote the sets of normal and anomalous feature locations in the contaminated training data, respectively.
Their sampling probabilities are defined as
\begin{equation}
\mathbb{P}(\Omega_{\mathcal{N}})
=
\frac{|\Omega_{\mathcal{N}}|}
{|\Omega_{\mathcal{N}}|+|\Omega_{\mathcal{A}}|},
\qquad
\mathbb{P}(\Omega_{\mathcal{A}})
=
\frac{|\Omega_{\mathcal{A}}|}
{|\Omega_{\mathcal{N}}|+|\Omega_{\mathcal{A}}|}.
\end{equation}

The expected RD objective can then be decomposed into the reconstruction losses over normal and anomalous feature locations:
\begin{equation}
\begin{split}
\mathcal{L}
=&\
\mathbb{P}(\Omega_{\mathcal{N}})
\cdot
\mathbb{E}_{u_i\sim\Omega_{\mathcal{N}}}
\left[
\ell_{\cos}
\left(
\mathcal{F}_{\mathcal{T},i},
\mathcal{F}_{\mathcal{S},i}
\right)
\right]
\\
&+
\mathbb{P}(\Omega_{\mathcal{A}})
\cdot
\mathbb{E}_{u_j\sim\Omega_{\mathcal{A}}}
\left[
\ell_{\cos}
\left(
\mathcal{F}_{\mathcal{T},j},
\mathcal{F}_{\mathcal{S},j}
\right)
\right],
\end{split}
\label{eq:supp_erm}
\end{equation}
where $u_i$ and $u_j$ index spatial feature locations.

Although the training set may contain a considerable proportion of anomalous images, anomalous regions generally occupy only a small fraction of their spatial locations. Therefore, at the feature-location level, we typically have
$\mathbb{P}(\Omega_{\mathcal{N}})
\gg
\mathbb{P}(\Omega_{\mathcal{A}})$.
Accordingly, the gradient with respect to the student parameters can be expressed as
\begin{equation}
\begin{split}
\frac{\partial \mathcal{L}}
{\partial \theta_{\mathcal{S}}}
=&\
\underbrace{
\mathbb{P}(\Omega_{\mathcal{N}})
\cdot
\mathbb{E}_{u_i\sim\Omega_{\mathcal{N}}}
\left[
\frac{\partial \ell_{\cos}}
{\partial \theta_{\mathcal{S}}}
\right]
}_{\text{Dominant Term}}
\\
&+
\underbrace{
\mathbb{P}(\Omega_{\mathcal{A}})
\cdot
\mathbb{E}_{u_j\sim\Omega_{\mathcal{A}}}
\left[
\frac{\partial \ell_{\cos}}
{\partial \theta_{\mathcal{S}}}
\right]
}_{\text{Subordinate Term}}.
\end{split}
\label{eq:supp_grad}
\end{equation}

Normal regions not only dominate numerically but also recur across different samples, providing relatively consistent reconstruction signals.
In contrast, anomalous regions vary substantially in location, scale, shape, and appearance, leading to less consistent gradient directions across samples.
Consequently, the aggregated optimization direction is primarily governed by normal reconstruction, allowing the student to learn stable reconstruction of shared normal structures while making the reconstruction of sparse anomalous features less stable and more sample-dependent.
This explains the Normal-Dominant Convergence described in Assumption~1.

\subsection{Limited Cross-Sample Anomaly Generalization}
\label{sec:supp_a2}

\textbf{Analysis.}
We analyze this assumption from the perspective of anomaly-specific reconstruction signals.
At an anomalous feature location, the teacher feature can be conceptually decomposed into a shared normal-related component and a sample-specific anomalous deviation:
\begin{equation}
\mathcal{F}_{\mathcal{T},j}
=
\mathcal{F}_{\mathcal{N},j}
+
\Delta_j,
\label{eq:supp_feature_decomposition}
\end{equation}
where $\mathcal{F}_{\mathcal{N},j}$ denotes the normal-related component and $\Delta_j$ denotes the anomalous deviation specific to sample $I_j$.

For an anomalous sample observed by the division-specific student $\mathcal{S}_k$, the reconstruction objective directly minimizes the discrepancy between the student output and $\mathcal{F}_{\mathcal{N},j}+\Delta_j$.
The student may therefore memorize the sample-specific deviation $\Delta_j$ and reconstruct the observed anomaly with a relatively small teacher--student discrepancy.

For an anomalous sample $I_j$ assigned to another division, however, $\mathcal{S}_k$ has never received direct supervision for its specific deviation $\Delta_j$. Its anomaly-related reconstruction behavior is instead determined by the diverse anomalous deviations seen in its own training division, which can be conceptually expressed as
\begin{equation}
\widehat{\Delta}_{k,j}
=
\sum_{I_i\in\mathcal{D}_k\cap\mathcal{A}_{\mathrm{train}}}
\alpha_{ij}\Delta_i,
\label{eq:supp_transferred_deviation}
\end{equation}
where $\alpha_{ij}$ conceptually characterizes the influence of the anomaly-specific reconstruction behavior learned from sample $I_i$ on the reconstruction of the target sample $I_j$.

Since the anomaly-specific deviations seen during training are generally diverse and lack a consistent direction, their transferred contributions are less likely to form a stable anomaly-related reconstruction signal for an unseen sample.

Consequently, the reconstruction of the unseen anomaly is dominated by the consistently learned normal-related component:
\begin{equation}
\mathcal{F}_{\mathcal{S}_k,j}
\approx
\mathcal{F}_{\mathcal{N},j}
+
\widehat{\Delta}_{k,j}
\approx
\mathcal{F}_{\mathcal{N},j},
\label{eq:supp_normal_like_reconstruction}
\end{equation}
which is consistent with the normal-like reconstruction observed in Fig.~3.

\IEEEpubidadjcol

This analysis is based on the empirical observation that
anomaly-specific deviations are generally less consistent across samples
than shared normal structures. Even when anomalous samples are
semantically related, their abnormal regions may still differ in
appearance, location, shape, scale, or local context. Therefore,
semantic similarity does not necessarily imply that the
sample-specific reconstruction behavior learned from one anomalous
sample can be fully transferred to another anomalous sample.

\subsection{Connection to Cross-Division Distillation}
\label{sec:supp_rationale_connection}

The first analysis explains why sufficient normal support enables each division-specific student to maintain stable normal reconstruction.
The second analysis explains why a student trained on another division is less likely to retain the anomaly-specific component of a target sample that it has not directly seen.
Therefore, the cross-division outputs can serve as candidate pseudo-normal features, which are further selected through Cross-Division Knowledge Aggregation to supervise the global student.

\section{Additional Ablation and Sensitivity Analyses}
\label{sec:supp_ablation}

Unless otherwise specified, all experiments in this section are conducted using the RD-based implementation CDD$^*$ on MVTec AD~\cite{bergmann2019mvtec}. The experimental configurations are consistent with those used in the main manuscript.

\subsection{Extended Analysis of the Number of Divisions}
\label{sec:supp_k}

\begin{figure}[t]
    \centering
    \includegraphics[width=\columnwidth]{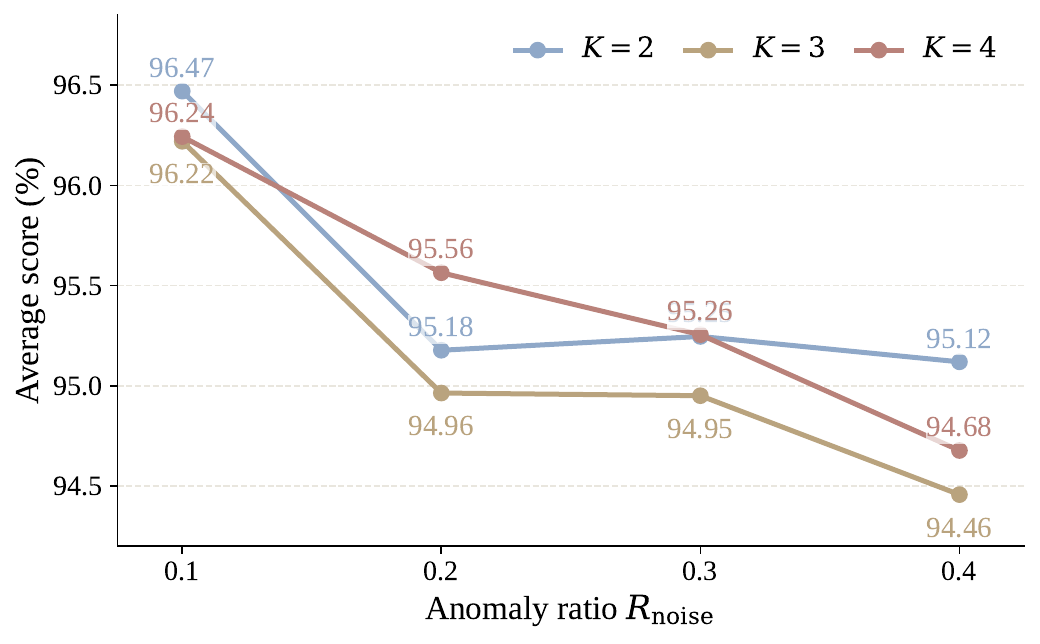}
    \caption{
    Sensitivity of the fixed division number $K$ under different
    contamination ratios $R_{\mathrm{noise}}$ on MVTec
    AD~\cite{bergmann2019mvtec}. The reported score is the arithmetic
    mean of I-AUC, P-AUC, and PRO.
    }
    \label{fig:supp_k_noise}
\end{figure}

To examine whether the number of divisions should be adjusted according to the contamination ratio, we further compare fixed $K\in\{2,3,4\}$ under different $R_{\mathrm{noise}}$ values on MVTec AD~\cite{bergmann2019mvtec}. The results are shown in Fig.~\ref{fig:supp_k_noise}.

The optimal fixed \(K\) does not increase monotonically with \(R_{\text{noise}}\), and \(K=2\) remains the best setting at \(R_{\text{noise}}=0.4\).  A larger \(K\) assigns fewer low-confidence samples to each division, which may facilitate the separation of anomalous samples across different divisions. However, it also reduces the amount of training data available to each division-specific student. Consequently, the potential benefit of finer division can be offset by insufficient training data for learning normal reconstruction. The final performance therefore reflects a trade-off between anomaly separation and training-data sufficiency, rather than being determined solely by the expected anomaly ratio within each division. 
Since CDD reconstructs the confidence-guided divisions at every epoch rather than relying on a one-time random partition, small \(K\) values remain robust across different anomaly ratios.

\subsection{Sensitivity to Maximum Feature Perturbation Strength}
\label{sec:supp_sigma}

\begin{figure}[t]
    \centering
    \includegraphics[width=\columnwidth]{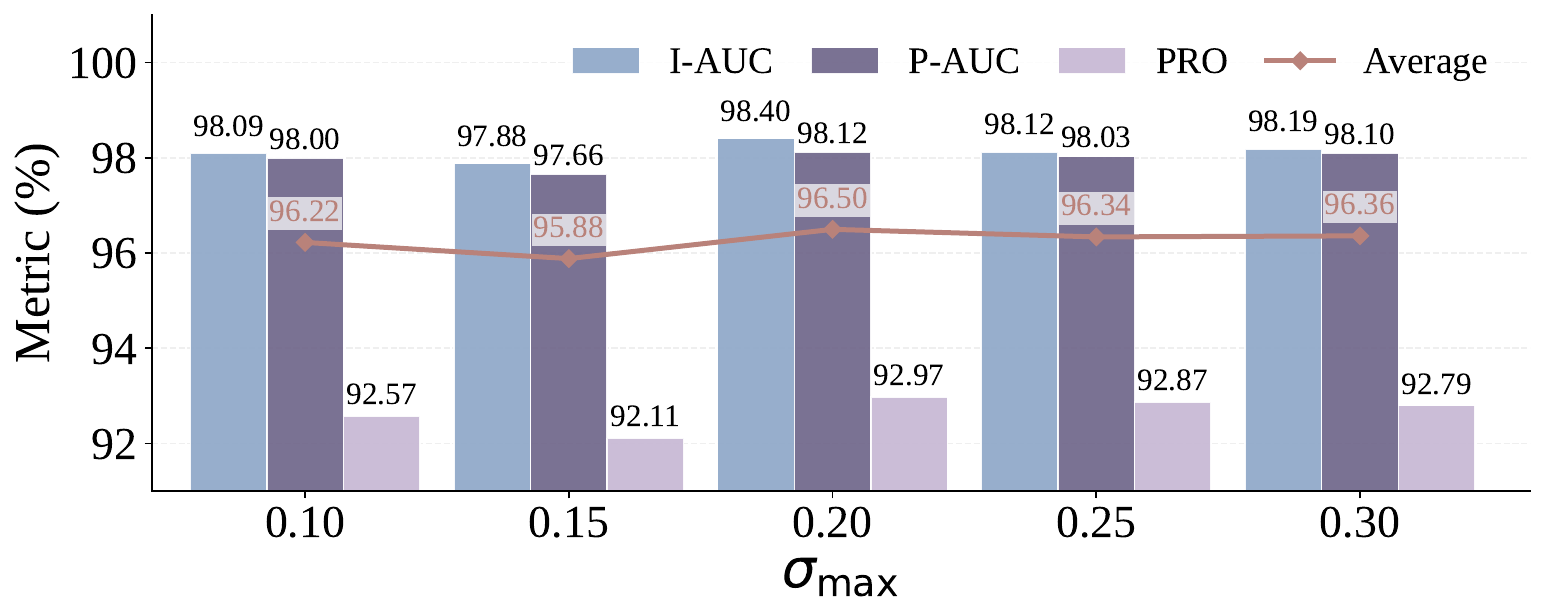}
    \caption{
    Sensitivity to the maximum Gaussian perturbation strength
    $\sigma_{\max}$ on MVTec AD-noise-0.1~\cite{bergmann2019mvtec}.
    The bars report I-AUC, P-AUC, and PRO, while the line denotes their
    arithmetic mean.
    }
    \label{fig:supp_sigma}
\end{figure}

To investigate the sensitivity of Gaussian feature perturbation, we vary the maximum perturbation strength as \(\sigma_{\max}\in\{0.10,0.15,0.20,0.25,0.30\}\), while keeping all other configurations unchanged. For each feature level, the actual perturbation standard deviation is randomly sampled from \(\mathcal{U}(0,\sigma_{\max})\).
The results in Fig.~\ref{fig:supp_sigma} indicate that CDD maintains stable performance across the tested range and is relatively insensitive to the exact perturbation upper bound within a moderate range. Among the evaluated values, \(\sigma_{\max}=0.20\) achieves the best overall performance, suggesting that a moderate perturbation strength provides an effective balance between feature-space regularization and representation preservation. These observations support \(\sigma_{\max}=0.20\) as a reasonable default for CDD.

\subsection{Sensitivity to the Regularization Schedule Parameter}
\label{sec:supp_p}

The parameter $p$ controls the shape of the S-shaped regularization coefficient $\lambda(e)$ used in Division-Specific Distillation. To evaluate its sensitivity, we vary $p\in\{0.5,1,2,4,6,8\}$ on MVTec AD-noise-0.1~\cite{bergmann2019mvtec} while keeping all other experimental configurations unchanged.

\begin{table}[t]
    \centering
    \caption{
    Sensitivity to the regularization schedule parameter $p$ on MVTec AD-noise-0.1~\cite{bergmann2019mvtec}. 
    }
    \label{tab:supp_p}
    \resizebox{0.8\columnwidth}{!}{
    \begin{tabular}{c|cccc}
        \toprule
        $p$ & I-AUC & P-AUC & PRO & Average \\
        \midrule
        0.5 & 0.9793 & 0.9800 & 0.9265 & 0.9619 \\
        1   & 0.9833 & 0.9797 & 0.9243 & 0.9624 \\
        2   & 0.9691 & 0.9787 & 0.9247 & 0.9575 \\
      \rowcolor{blue!10}  4   & 0.9840
            & 0.9812
            & 0.9297
            & 0.9650 \\
        6   & 0.9841
            & 0.9805
            & 0.9271
            & 0.9639 \\
        8   & 0.9726 & 0.9782 & 0.9259 & 0.9589 \\
        \bottomrule
    \end{tabular}}
\end{table}

As shown in Table~\ref{tab:supp_p}, the performance remains relatively stable over a broad range of $p$, with the average score varying from 0.9575 to 0.9650 and exhibiting no monotonic trend. The final setting $p=4$ achieves the best average performance, as well as the best P-AUC and PRO. Although $p=6$ obtains a marginally higher I-AUC, its average performance is lower than that of $p=4$.
These results indicate that the performance does not depend on a narrowly selected value of $p$, and we therefore take $p=4$ as the default configuration.

\subsection{Training Dynamics of Confidence-Guided Division Construction}
\label{sec:supp_cgdc_dynamics}

To examine the behavior of Confidence-Guided Division Construction when the confidence estimates of the global student are still uncertain, we trace the anomaly-to-normal ratios of both the constructed divisions and the High-Confidence (HC) division throughout training. Ground-truth anomaly labels are used only for this post-hoc diagnostic analysis and are not involved in model training.

\begin{figure*}[t]
    \centering
    \includegraphics[width=0.85\textwidth]{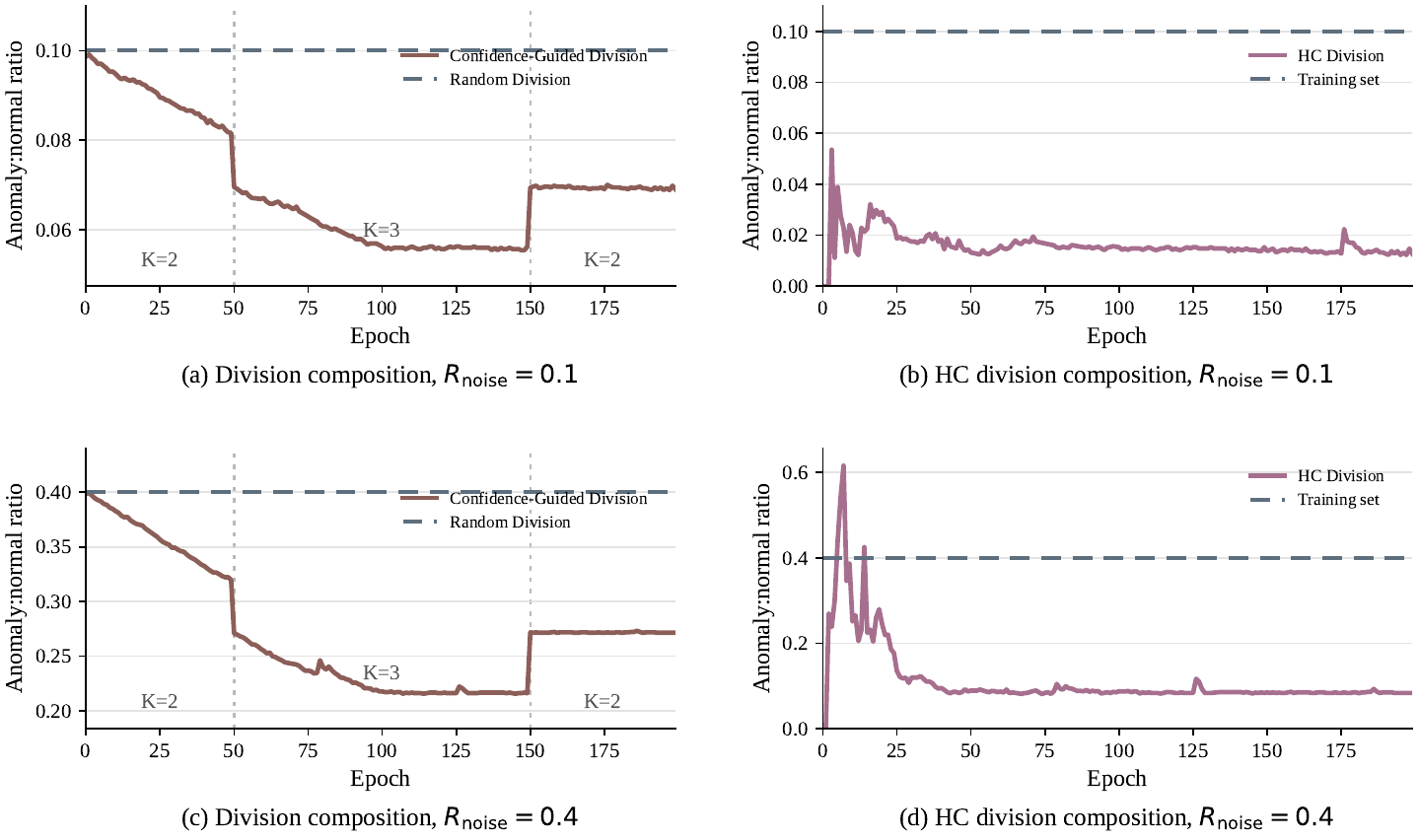}
    \caption{
    Training dynamics of Confidence-Guided Division Construction on MVTec AD~\cite{bergmann2019mvtec}.
    (a, c) Average anomaly-to-normal ratio of the divisions constructed with confidence guidance, compared with the ratio obtained by pure random division, on MVTec AD-noise-0.1 and MVTec AD-noise-0.4, respectively.
    (b, d) Anomaly-to-normal ratio of the High-Confidence (HC) division \(\mathcal{D}^{HC}\), compared with that of the complete training set, on MVTec AD-noise-0.1 and MVTec AD-noise-0.4, respectively.
    }
    \label{fig:supp_hc_dynamics}
\end{figure*}

\begin{figure}
  \centering
  \includegraphics[width=0.7\linewidth]{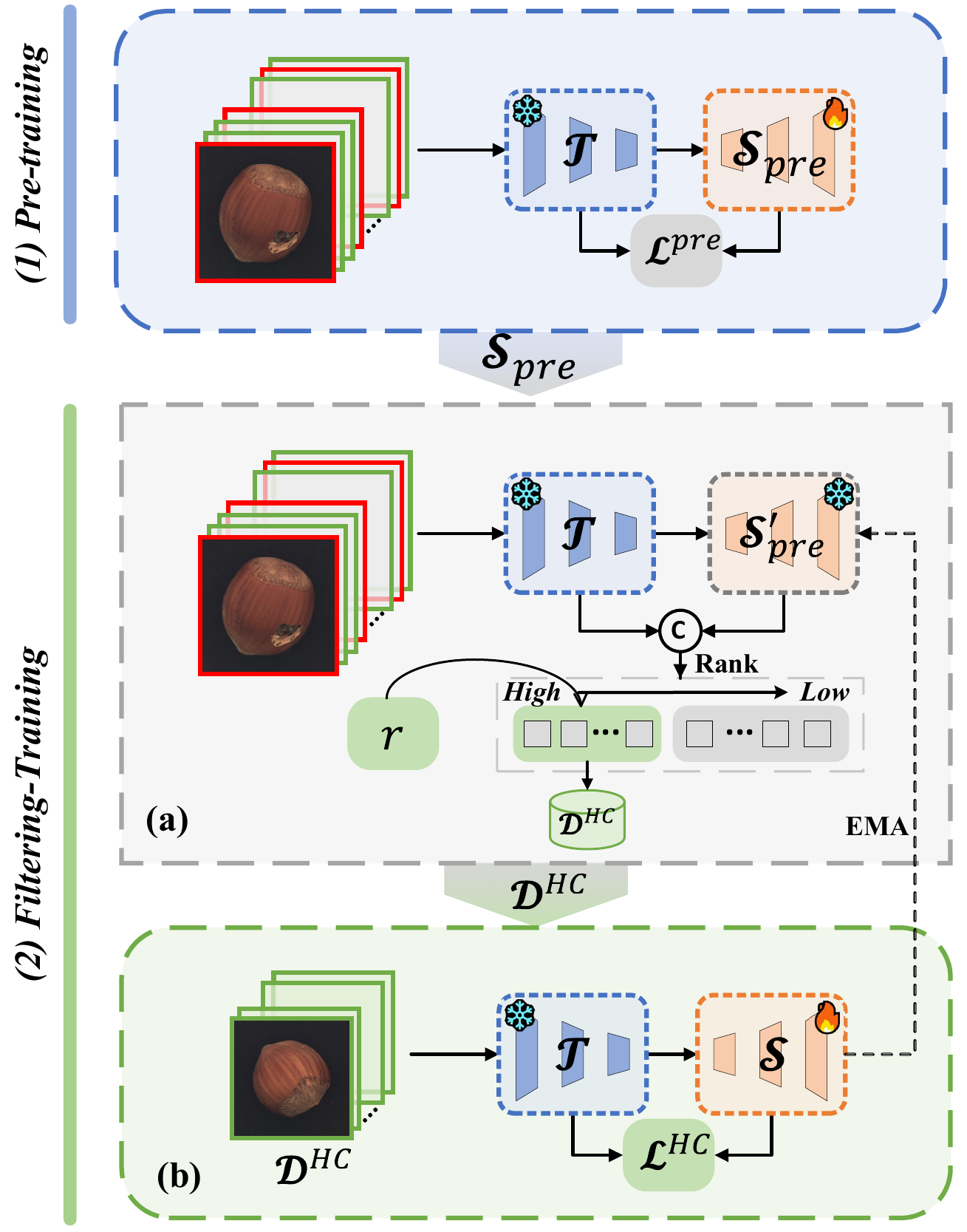}
  \caption{Pipeline of RD w/ Filter. The process consists of two phases: (1) Pre-training Phase: A standard RD student \(\mathcal{S}_{pre}\) is trained on the entire contaminated dataset. (2) Filtering-Training Phase: This iterative phase runs for multiple epochs, each containing (a) Filtering Step: The model \(\mathcal{S}'_{pre}\) selects the top $r$ ($r=50\%$) most normal samples as the high-confidence set \(\mathcal{D}^{HC}\); (b) Training Step: The student \(\mathcal{S}\) is trained on \(\mathcal{D}^{HC}\). These two steps are repeated iteratively, with \(\mathcal{S}'_{pre}\) being updated via EMA from \(\mathcal{S}\) after each epoch.
  }
  \label{fig:rd_filter_pipeline}
\end{figure}

\begin{algorithm}[t]
\caption{RD w/ Filter} \label{alg:rd_filter}
\begin{algorithmic}[1]
\State \textbf{Input:}  \(\mathcal{I}_{train} = \{I_i\}_{i=1}^{N}\),  \(\mathcal{T}\)
\State \textbf{Output:} \(\mathcal{S}_{final}\)
\State \textbf{Parameters:} Pre-training epochs \(E_{pre}=200\), filtering-training epochs \(E_{filter}=200\), filtering ratio \(r=0.5\), EMA decay \(\alpha=0.99\)

\State \textbf{\textit{// Phase I: Pre-training}}
\State Initialize student \(\mathcal{S}_{pre}\)
\For{\(e = 0\) to \(E_{pre}-1\)}
    \State Train \(\mathcal{S}_{pre}\) on all samples in \(\mathcal{I}_{train}\) using \( \mathcal{L}^{pre}\)
\EndFor

\State \textbf{\textit{// Phase II: Filtering-Training}}
\State Initialize student \(\mathcal{S} \gets \mathcal{S}_{pre}\)  \Comment{Training network}
\State Initialize scoring model \(\mathcal{S}^{'}_{pre} \gets \mathcal{S}_{pre}\) \Comment{For confidence scoring}

\For{\(e = 0\) to \(E_{filter}-1\)}
    \State \textbf{\textit{// Filtering Step}}
    \State Use \(\mathcal{S}^{'}_{pre}\) to compute confidence scores for all samples in \(\mathcal{I}_{train}\)
    \State Select top \(r \cdot N\) samples with highest scores as \(\mathcal{D}^{HC}\)
    
    \State \textbf{\textit{// Training Step}}
    \State Train \(\mathcal{S}\) for one epoch on \(\mathcal{D}^{HC}\) using RD loss
    
    \State Update \(\mathcal{S}'\) parameters by EMA: \(\theta_{\mathcal{S}^{'}_{pre}} \gets \alpha \theta_{\mathcal{S}^{'}_{pre}} + (1-\alpha) \theta_{\mathcal{S}}\)
\EndFor

\State \(\mathcal{S}_{final} \gets \mathcal{S}\)
\State \Return \(\mathcal{S}_{final}\)
\end{algorithmic}
\end{algorithm}

As shown in Fig.~\ref{fig:supp_hc_dynamics} (a) and (c), the constructeddivisions are equivalent to pure random division at initialization since \(r(0)=0\). Their anomaly-to-normal ratio subsequently decreases and becomes clearly lower than the random-division reference. This difference is maintained during the remaining training epochs. The abrupt changes at Epochs~50 and~150 correspond to the predefined changes in the number of divisions under \(K=\{2,3,3,2\}\), rather than directly reflecting abrupt changes in the confidence estimates.

Fig.~\ref{fig:supp_hc_dynamics} (b) and (d) further show that, despite fluctuations during the earliest epochs, the anomaly-to-normal ratio of \(\mathcal{D}^{HC}\) rapidly decreases and subsequently remains substantially lower than that of the complete training set. This suggests that early confidence uncertainty does not develop into progressively increasing contamination of the HC division.

\section{Implementation Details of RD w/ Filter}
\label{app_filter}

As introduced in Section I, a straightforward strategy for the FUAD task is to filter out suspected anomalies, thereby converting it into a standard UAD problem. To rigorously benchmark against this idea, we implement a filtering baseline based on Reverse Distillation (RD)~\cite{deng2022anomaly} framework, denoted as \textbf{RD w/ Filter}. Its implementation, visually summarized in Fig.~\ref{fig:rd_filter_pipeline} and formally outlined in Algorithm~\ref{alg:rd_filter}, consists of two phases: a \textbf{Pre-training Phase} and a \textbf{Filtering-Training Phase}.

\paragraph{Pre-training Phase.}
A standard RD model is first pre-trained on the entire contaminated training set for 200 epochs. This provides a preliminary model \( \mathcal{S}_{pre} \) that, while potentially overfitting to some anomalies, can serve as a reasonable initial anomaly scorer. The model is trained using the standard cosine distance loss 
\( \mathcal{L}^{pre} = \mathbb{E}_{I_i \sim \mathcal{I}_{train}} \ell_{cos}(\mathcal{F}_{\mathcal{T},i}, \mathcal{F}_{\mathcal{S}_{pre},i}) \).

\paragraph{Filtering-Training Phase.}
This phase runs for another 200 epochs and is performed iteratively, with one filtering step followed by one training step per epoch.

\begin{itemize}
    \item \textbf{Filtering Step:} At the beginning of each epoch, we use the current student model \( \mathcal{S}^{'}_{pre} \) (whose parameters are initialized from \( \mathcal{S}_{pre} \)) to compute a confidence score \(\mathrm{Conf}_i\) (higher scores are more likely to indicate normality) for each training sample \(I_i \in \mathcal{I}_{train}\). (The score is calculated in the same way as in Section IV-A.) The scores are sorted, and only the top $r$ ($r=0.5$) of samples with the lowest scores (called the high-confidence normal set \( \mathcal{D}^{HC} \)) are retained for that epoch's subsequent training.
    \item \textbf{Training Step:} The student model \( \mathcal{S} \) (also initialized from \( \mathcal{S}_{pre} \)) is then trained for one epoch exclusively on \( \mathcal{D}^{HC} \), minimizing the loss \( \mathcal{L}^{HC} = \mathbb{E}_{I_i \sim \mathcal{D}^{HC}} \ell_{cos}(\mathcal{F}_{\mathcal{T},i}, \mathcal{F}_{\mathcal{S},i}) \).
\end{itemize}

Crucially, the model \( \mathcal{S}'_{pre} \) used for scoring is updated via an Exponential Moving Average (EMA)~\cite{tarvainen2017mean} of the trained student \( \mathcal{S} \) after each epoch. This ensures that the filtering mechanism evolves and improves as the student becomes more proficient.

\end{document}